\documentclass[conference]{IEEEtran}
\IEEEoverridecommandlockouts
\usepackage{cite}
\usepackage{amsmath,amssymb,amsfonts}
\usepackage{graphicx}
\usepackage{textcomp}
\usepackage{xcolor}
\usepackage[font={footnotesize}]{caption}

\usepackage[para,online,flushleft]{threeparttable}

\usepackage{hyperref}
\usepackage{url}

\usepackage{algpseudocode}

\usepackage{booktabs}
\usepackage{adjustbox}
\usepackage{subfig}
\usepackage{enumitem}
\usepackage[ruled,vlined]{algorithm2e}
\usepackage[ruled]{algorithm2e}
\usepackage{multirow}
\usepackage{marvosym}
\usepackage{dblfloatfix} 

\usepackage{wrapfig}

\SetKwInput{kwSym}{Notation}
\SetKwInput{kwCalc}{Calculate range of feature embeddings}
\SetKwInput{kwSel}{Extract FE for each training set instance}
\SetKwInput{kwLab}{Label each instance as Safe, Border, Rare, Outlier}

\def\BibTeX{{\rm B\kern-.05em{\sc i\kern-.025em b}\kern-.08em
    T\kern-.1667em\lower.7ex\hbox{E}\kern-.125emX}}
    
  \makeatletter
    \newcommand{\linebreakand}{%
      \end{@IEEEauthorhalign}
      \hfill\mbox{}\par
      \mbox{}\hfill\begin{@IEEEauthorhalign}
    }
    \makeatother

\begin{document}
\title{Interpretable ML for Imbalanced Data}

\author{\IEEEauthorblockN{Damien A. Dablain}
\IEEEauthorblockA{\textit{Lucy Family Institute for Data \& Society} \\ \textit{Dept. Computer Science and Engineering} \\
\textit{University of Notre Dame}\\
Notre Dame, IN 46556, USA \\
ddablain@nd.edu}
\and
\IEEEauthorblockN{Colin Bellinger}
\IEEEauthorblockA{\textit{National Research Council of Canada}\\
 Ottawa, Canada, K1A 0R6\\
colin.bellinger@nrc-cnrc.gc.ca}
\linebreakand
\IEEEauthorblockN{Bartosz Krawczyk}
\IEEEauthorblockA{\textit{Dept. Computer Science} \\
\textit{Virginia Commonwealth University}\\
Richmond, VA 23824, USA \\
bkrawczyk@vcu.edu}
\and
\IEEEauthorblockN{David W. Aha}
\IEEEauthorblockA{\textit{Navy Center for Applied Research in AI,}\\ {Naval  Research Laboratory}\\
 Washington, DC 20375\\
david.aha@nrl.navy.mil}

\and
\IEEEauthorblockN{Nitesh V. Chawla}
\IEEEauthorblockA{\textit{Lucy Family Institute for Data \& Society} \\ \textit{Dept. Computer Science and Engineering} \\
\textit{University of Notre Dame}\\
Notre Dame, IN 46556, USA \\
nchawla@nd.edu}

}

\maketitle

\begin{abstract}
Deep learning models are being increasingly applied to imbalanced data in high stakes fields such as medicine, autonomous driving, and intelligence analysis. Imbalanced data compounds the black-box nature of deep networks because the relationships between classes may be highly skewed and unclear. This can reduce trust by model users and hamper the progress of developers of imbalanced learning algorithms. Existing methods that investigate imbalanced data complexity are geared toward binary classification, shallow learning models and low dimensional data. In addition, current eXplainable Artificial Intelligence (XAI) techniques mainly focus on converting opaque deep learning models into simpler models (e.g., decision trees) or mapping predictions for specific \textit{instances} to inputs, instead of examining \textit{global} data properties and complexities. Therefore, there is a need for a framework that is tailored to modern deep networks, that incorporates large, high dimensional, multi-class datasets, and uncovers data complexities commonly found in imbalanced data (e.g., class overlap, sub-concepts, and outlier instances). We propose a set of techniques that can be used by both deep learning model users to identify, visualize and understand class prototypes, sub-concepts and outlier instances; and by imbalanced learning algorithm developers to detect features and class exemplars that are key to model performance. Our framework also identifies instances that reside on the \textit{border} of class decision boundaries, which can carry highly discriminative information. Unlike many existing XAI techniques which map model decisions to gray-scale pixel locations, we use saliency through back-propagation to identify and \textit{aggregate} image color bands across \textit{entire classes}. This facilitates the high-level analysis of foreground and  background colors used by models to discern classes (e.g., blue sky for airplanes). Our framework is publicly available at \url{https://github.com/dd1github/XAI_for_Imbalanced_Learning}

\end{abstract}



\maketitle

\section{Introduction}

Deep learning models, including convolutional neural networks (CNNs), are increasingly being used in high-stakes fields such as medical diagnosis \cite{tjoa2020survey}, autonomous driving \cite{levinson2011towards,huang2018apolloscape}, criminal justice \cite{wexler2017computer}, environmental remediation \cite{mcgough2018bad} and intelligence analysis \cite{dorton2022naturalistic}. Yet, their decisions can be opaque, which reduces the trust of model users and makes it challenging for machine learning (ML) algorithm developers to diagnose and improve model performance. The black-box nature of neural networks has spawned the field of eXplainable Artificial Intelligence (XAI), which seeks to develop techniques to interpret and explain models to increase trust, verifiability, and accountability \cite{gunning2019darpa}. Since no clear, commonly agreed upon definition of \textit{explanation} and \textit{interpretation} exists \cite{linardatos2020explainable}, the terms XAI and Interpretable Machine Learning (IML) are often used interchangeably throughout the literature. For purposes of this paper, we use the term "IML" to distinguish it from methods that offer an "explanation", which has a rich history in the social sciences as involving human interaction and human subject studies to evaluate the quality of an explanation \cite{miller2019explanation,hoffman2018metrics}.

The perceived need to enhance the interpretability of deep learning models has resulted in a number of techniques that are specifically targeted to fields that use ML, such as medicine \cite{bruckert2020next,han2020augmented}, air traffic control \cite{xie2021explanation},  finance \cite{chen2018interpretable}, and autonomous driving \cite{kim2017interpretable}. However, there is a paucity of IML techniques that have been explicitly adapted for imbalanced learning.

In some real-world situations, CNNs may be trained with imbalanced data \cite{johnson2019survey,liu2019large,tang2020long}, such that there is a difference in the number of training examples for one or more classes. At inference time, deep learning models that incorporate imbalanced learning algorithms typically offer users single instance predictions or a macro assessment of model accuracy. These single metrics for deep models typically consist of balanced accuracy, macro F1 measure, or geometric mean. All of these measures use a single scalar to describe the performance of the model on \textit{all} classes. We believe that these measures fail to adequately explain the complex interplay of class overlap, sub-concepts and the role of specific input and latent features on a deep network's decision process.

In some cases, shallow learning for imbalanced data has incorporated additional measures of global data complexity \cite{barella2021assessing}; however, these measures have largely been applied at the \textit{input} data level. Shallow machine learning models also traditionally dealt with smaller scale datasets and a compact feature space. In contrast, deep models are trained on datasets with tens or hundreds of thousands of examples and extremely large input feature spaces, especially in computer vision. To better understand how deep networks act on imbalanced data, we incorporate data complexity visualizations at both the input and latent feature level.

Interpretation is critical to both imbalanced learning and IML; although both fields have approached it from different perspectives. IML has generally focused on \textit{model} interpretability; whereas imbalanced learning has sought to better understand \textit{data} complexity. When applied to deep learning, IML methods are generally designed to explain a model's internal representations, connect outputs with inputs, and use more transparent models to explain complicated neural networks \cite{gilpin2018explaining,murdoch2019definitions}. In contrast, imbalanced learning has typically sought to understand the interplay of class imbalance with overlap, sub-concepts and data outliers. In addition, many IML techniques usually seek to explain model decisions with respect to specific \textit{instances}; whereas imbalanced learning is generally concerned with the global properties of \textit{entire classes}. 

In this work, we combine facets of both fields into a single framework to better understand a CNN's predictions with respect to imbalanced data. We do not develop a single method to improve the interpretability of complex, imbalanced datasets. Rather, we propose a framework and suite of tools that can be used by both model developers and users to better understand imbalanced data and how a deep network acts on it.

We combine both approaches because we use the internal representations learned by a trained deep convolutional neural network to explore data complexity. We extract and aggregate the CNN's learned, latent feature embeddings to better understand data complexity and how a model uses its latent features to make class decisions. For example, traditional instance selection for resampling in imbalanced learning has been performed as a pre-processing step on input data; however, we consider instance selection based on a model's latent, lower dimensional representation of data inputs. 

We also suggest potential future research directions for imbalanced learning model developers based on insights drawn from our framework. We use frequent visualizations of the complex data relationships and features learned by a model, in the spirit of many IML methods; however, we apply these visualization at a more global (class) level instead of explaining specific instances.

Our goal is to provide a suite of tools to facilitate understanding of data complexity for imbalanced learning so that model users can better diagnose the underlying problems that contribute to model under-performance. By improving our understanding of how CNN's work on complex data, we can develop insights into some of the questions that are central to imbalanced learning:
\begin{itemize}
    \item \textbf{Which class instances are important for sampling purposes?} Three of the key methods used to address data imbalance - oversampling, undersampling and ensembles - all rely on careful instance selection. By examining a CNN's latent representation of class features, we partition the instances of each class into safe (homogeneous), border (near decision boundary), rare (sub-concepts) and outlier (potential noise) categories, which can be used to improve sample selection.
    \item \textbf{How should novel cost-sensitive loss functions be identified?} Many of the existing cost-sensitive algorithms used in imbalanced learning assign a higher cost based on minority class \textit{instances}. An under-explored avenue within cost-sensitive learning could involve assigning costs based on class-specific \textit{features}. We explore and visualize a CNN's latent representations, which are derived from its feature maps. These feature embeddings, which are indicators of class overlap (and model feature entanglement) can potentially serve as the basis for novel loss functions applied on the \textit{latent feature} versus instance level.
\end{itemize}

In this paper, we make the following research contributions to the field of imbalanced learning:
\begin{itemize}
    \item \textbf{Framework for understanding the complexity of imbalanced data used in deep networks.} Existing deep learning methods trained with imbalanced data typically use a single metric to describe model performance on all classes. We provide a suite of visualizations so that model users and algorithm developers can better understand dataset specific concepts that are central to imbalanced learning (e.g., class overlap, sub-concepts, prototypes, relative importance of input and latent features). 
    \item \textbf{Extensible to large, high dimensional, multi-class datasets commonly deployed in deep learning.} Many existing imbalanced learning techniques that assess data complexity are designed for binary classification on datasets with only a few thousand instances and less than a hundred features (e.g., shallow learning). Because we use the low-dimensional latent representations learned by a CNN in our framework, we are able to efficiently visualize the key data properties of majority and minority classes on high dimensional, multi-class datasets with a large number of instances.
    \item \textbf{Predict relative validation set false positives by class with training data}. In many fields (e.g., chemistry and network anomaly detection), minority class data is precious and carving out training data for validation purposes can be costly. We show that the likely classes that will produce the most validation set false positives for a given reference class can be predicted \textit{solely} from training data. This insight can save valuable data which would have been used in the validation process instead for training purposes.
    \item \textbf{Class saliency color visualizations.}  Existing IML methods display black and white heatmaps of pixel saliency for single dataset instances. We, instead, visualize the most salient colors used by CNN models to identify entire classes. Similar to IML saliency methods, we use the gradient of individual instances to map decisions to input pixels; however, we aggregate this information efficiently across all instances in large datasets by using color prototypes and latent feature embeddings. 
    \item \textbf{Specific insights tailored to imbalanced learning.} For each component of our framework, we offer examples of specific insights that can be gleaned from our IML techniques and also suggest future research directions based on our framework. These insights are relevant to both model developers (to improve deep learning algorithms) and users (to better understand model decisions). 
\end{itemize}

\section{Background \& Related Work}\label{sec:RW}

In this section, we introduce the guiding principles in IML and imbalanced learning that animate our framework. The related work is organized around the following key concepts:
\begin{itemize}
    \item \textbf{Data is an important element of model understanding.} Advances in deep learning have been built, in part, on access to large amounts of data. Therefore, it is critical to understand how the model organizes data into low dimensional representations used for classification.
    \item \textbf{Need for global data complexity insights to explain deep networks.} Many current IML methods are instance-specific; whereas imbalanced learning explanation requires intuition about global (class) characteristics. 
    \item \textbf{CNN texture bias as interpretation.} The perceived texture bias of CNNs can be used to extract informative global, class-wise insights.
\end{itemize}
We also discuss the prior work that inspires our research and how our approach differs from previous methods. 

\textbf{Centrality of data to deep learning \& class imbalance understanding.} Deep learning has shown significant progress in the past decade due, in part, to the ubiquity of low cost and freely available data \cite{marcus2018deep}. Deep networks are typically trained on thousands and even millions of examples to minimize the average error on training data (empirical risk minimization) \cite{zhang2018mixup}. As the size and complexity of modern datasets grow, it is increasingly important to provide model users and developers vital information and visualizations of representative examples that carry interpretative value \cite{bien2011prototype}. In addition, when deep networks fail on imbalanced data, it is not always intuitive to diagnose the role of data complexity on classifier performance \cite{kabra2015understanding}.

In imbalanced learning, several studies have assessed the complexity of the data used to train machine learning models; however, many of these studies were developed for small scale datasets used in shallow models. Barella et al. \cite{barella2021assessing} provide measures to assess the  complexity of imbalanced data. Their package is written for binary classification and is based on datasets with 3,000 or fewer instances and less than 100 features. Batista et al. \cite{batista2004study} determined that complexity factors such as class overlap are compounded by data imbalance. Their study was performed with respect to binary classification on datasets with 20,000 or fewer examples and 60 or fewer features. Their conclusion that class overlap is a central problem when studying class imbalance was confirmed by  \cite{denil2010overlap,prati2004class,garcia2007empirical}. Rare instances, class sub-concepts and small disjuncts can also exacerbate data imbalance and add to data complexity \cite{jo2004class,weiss2004mining}. Similarly, outliers and noisy samples can contribute to classifier inaccuracy \cite{aha1992tolerating}.

Several studies have shown that it is not only class overlap, but the \textit{concentration} of instances and features in the area of the decision boundary that cause false positives \cite{smote-variants}. In imbalanced learning, this observation has motivated a number of resampling methods that seek to balance the density of class examples along the decision boundary \cite{han2005borderline,he2008adasyn,dablain2022efficient}. These methods are typically applied to balance the number of class \textit{instances}.

Therefore, understanding data complexity, including class overlap, rare, border and outlier instances, is critical to improving imbalanced learning classifiers. This is especially important in deep learning, where opaque models trained with batch processing may obscure underlying data complexity. Unlike prior work, which explained data complexity by examining model inputs, we explain data complexity via the latent features learned by a model. These low-dimensional representations are the raw material used by the final classification layer of CNNs to make their predictions.

\textbf{Global (class) vs. instance level interpretation.}
Several studies have shown that intepretation is critical to machine learning model user satisfaction and acceptance \cite{teach1981analysis,ye1995impact}. It is also important for model developers for diagnostic and algorithm improvement purposes. Explanation is central to both IML and imbalanced learning; however, these fields approach it in different ways.  

In IML, great strides have been made to increase model interpretability by describing the inner workings of models and justifying how or why a model developed its prediction (post-hoc explanation) \cite{kenny2021explaining}. However, no clear definition of explainability exists, nor are there authoritative metrics that measure whether a method improves model interpretability \cite{guidotti2018survey,doshi2017towards}; although Hoffman et al. \cite{hoffman2018metrics} have proposed several metrics, including: (1) mental model understanding, (2) human-system task performance, and (3) appropriate setting of trust. In general, IML techniques can roughly be divided into four groups.

First, there are methods that explain a model's predictions by attributing decisions to inputs, including pixel attribution through back-propagation \cite{simonyan2013deep,selvaraju2017grad,sundararajan2017axiomatic}. Several recent studies have questioned the accuracy of pixel attribution methods \cite{kindermans2019reliability,ghorbani2019interpretation}, although the method developed by Simonyan et al. was generally found to be reliable \cite{adebayo2018sanity}. All of these methods work on \textit{single} data instances and do not provide an overall view of class homogeneity, sub-concepts, or outliers. In fact, a recent study found that the vast majority of IML methods were instance based \cite{huber2021local}.

\begin{figure*}[!t]
   \vspace{-0.2cm}
  \centering
  \subfloat{\includegraphics[width=0.99\textwidth]{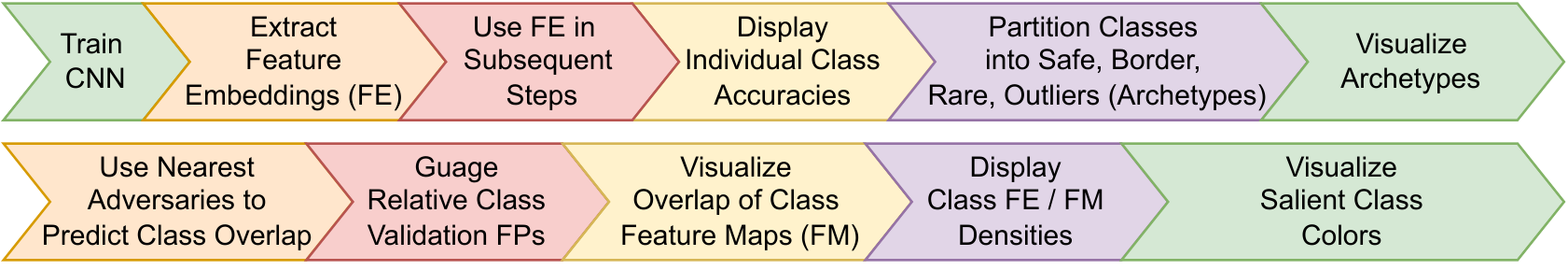}}
  \caption{This diagram outlines the main components of our IML  framework for imbalanced learning. It starts broadly by examining individual class accuracies and visualizing representative class archetypes.  Then, it visualizes class overlap at both the instance and latent feature level. Finally, it zooms-in on specific class latent feature densities and colors in pixel space, so that broader issues spotted in earlier stages can be examined in more detail at a class-of-interest level.}
  \label{fig_6_frame}
  \vspace{-0.4cm}
\end{figure*}

Second, explanations by example provide evidence of the model's prediction by citing or displaying similarly situated instances that produce a similar result or through counter-factuals - instances that are similar, yet produce an opposite or adversary result \cite{lipton2018mythos,keane2019case}. Both of these approaches are based on nearest neighbors, which can provide intuitive and plausible instances to explain a model's decision \cite{biran2017explanation,sormo2005explanation}. Like pixel attribution methods, this approach only provides explanations for single instances or predictions.

Third, there are methods that explain a complex neural network by replacing, or modifying, it with a simpler model. These approaches include local interpretable model explanations (LIME) \cite{ribeiro2016should}, Shapley values (occlusion based attribution) \cite{shapley1997value,sundararajan2020many}, and the incorporation of the K-nearest neighbor (KNN) \cite{fix1989discriminatory, cover1967nearest} algorithm into deep network layers \cite{papernot2018deep}. Both LIME and Shapley values can be computationally expensive because they involve repeated forward passes through a model \cite{achtibat2022towards}. 

Finally, there are IML methods that extract rules learned by a model \cite{zilke2016deepred} and the features or concepts represented by individual filters or neurons \cite{gilpin2018explaining}.

In summary, many existing IML methods offer interpretations for single instances and do not describe the broad class characteristics learned, or used, by a neural network to arrive at its decision. By contrast, in imbalanced learning, the focus of most explanatory methods has been on the global properties of data and classes within a dataset, including the inter-play of class imbalance and data complexity factors, such as class overlap, sub-concepts and noisy examples.

Napierala and Stefanowski partitioned \textit{minority} classes into instances that were homogeneous (safe), residing on the decision boundary (border), rare (aka sub-concepts and disjuncts), and outliers \cite{napierala2016types}. They used the KNN and Support Vector Machine (SVM) algorithms to separate \textit{input} data based on the local neighborhood (KNN) and support vectors. They hypothesized that separating classes into sub-groups provided insight into class overlap and sub-concepts. We extend their method to \textit{both} majority and minority classes and use a model's \textit{latent} representations to identify instance similarity based on the local neighborhood, instead of using the input space. In addition, a number of resampling methods in imbalanced learning have relied on border examples to provide discriminative class information \cite{han2005borderline,he2008adasyn}.

\textbf{CNN texture bias as explanation.} Humans are believed to strongly emphasize shape for purposes of object recognition  \cite{landau1988importance,soja1991ontological,hermann2022understanding}, which may explain why IML visualization techniques that presumably rely on shape have been used to explain CNN predictions \cite{zeiler2014visualizing}. Early studies of CNNs hypothesized that their impressive performance was due to their ability to mimic the shape bias of the human visual system \cite{kriegeskorte2015deep,lecun2015deep,kubilius2016deep}. However, recent work has challenged that assumption and has demonstrated that CNNs emphasize texture over shape for object recognition tasks \cite{geirhos2018imagenet,baker2018deep,hermann2020origins}. 

A precise definition of texture remains elusive \cite{haralick1979statistical}. Coggins \cite{coggins1983framework} compiled a list of texture definitions that included repetitive patterns of tonal primitives, and coarseness, fineness or bumpiness of an image region. 
Tuceryan and Jain \cite{tuceryan1993texture} defined it as a spatial variation in pixel intensities. Due to the difficulty of precisely defining texture, we focus on one of its properties - color or chromaticity of a region. We use a CNN's color bias as \textit{explanation}. As discussed in more detail in Section~\ref{sec:texture}, we combine both saliency maps and pixel aggregation to reveal the most prevalent colors that a CNN relies on to distinguish a class.

\section{IML Framework for Imbalanced Learning}\label{sec:frame}

In this section, we outline our framework for applying IML to imbalanced, complex data contained in large datasets with high dimensional input spaces. Our framework is built on feature embeddings (FE). It starts broadly by visualizing individual class accuracy and representative data sub-groups within classes (archetypes), which allows for a quick overview of large, complex datasets with just a few examples. Then, relative  overlap can be visualized across multiple classes in terms of both instances and specific latent features. Finally, our framework allows for zooming in on specific classes to view feature densities compared with overlapping classes, and colors that are most salient for a specific class of interest (e.g., the minority class). This format allows both model users and developers to gain broad intuition of multiple class complexities before focusing on specific classes of interest.

The basic components of the framework are graphically shown in Figure~\ref{fig_6_frame}. Each component is discussed below.

\subsection{Feature Embeddings: Low Dimensional, Latent Representations Learned by a CNN} 
For large image datasets, it may be computationally challenging to examine class characteristics in input space. For example, the INaturalist dataset contains over 500,000 images of size 3 X 600 X 800 (1.4 million pixels each). Including all of these images in working memory to compute pair-wise instance distances, or class properties, may be infeasible for most imbalanced learning model users or developers.

To make our analysis of imbalanced data complexity more tractable, we work with the low dimensional feature embeddings learned by a CNN. We select the latent representations in the penultimate layer of a CNN, after pooling, and before the fully connected classification layer. We select these features because: (1) they represent the output of the CNN's final feature maps after pooling, (2) they serve as direct input to the final, fully connected classification layer, and (3) the features learned by the last CNN layers are specific to the dataset and task \cite{yosinski2014transferable}. We refer to these features as \textit{feature embeddings (FE)}. Feature embeddings can be extracted from a trained CNN and used to analysze dataset complexity and to better understand how the model acts on data. All of the other components in our framework rely on FE.

We believe that a model's latent feature representations are more informative of dataset complexity than real space features. For a CNN, feature embeddings represent lower dimensional, invariant characteristics learned from the underlying data distribution. These embeddings are directly used by the model to classify images. Lower tier representations have been shown to capture only basic image properties which are not necessarily invariant \cite{yosinski2014transferable}.

\subsection{Individual Class Accuracy} This component of the framework consists of a bar chart of individual class accuracy, combined with the class imbalance ratio. We believe that displaying individual class accuracy, along with the imbalance level, is more informative than a single metric, which is applied at the dataset, instead of class, level.  This visualization provides a high-level view of model performance and the impact of imbalance on accuracy. 

In a single chart, it provides an indication whether imbalance is the main driver of accuracy, or if other factors, such as class overlap and data complexity, play a role. For example, if the imbalance ratio increases, but accuracy also increases (or vice versa), then other data complexity factors besides under-represented classes, may be affecting accuracy. See Section~\ref{sec:acc} for visualizations.

\subsection{Class Archetypes}
Once we acquire a broad sense of individual class accuracy, we divide each class into four sub-categories or archetypes: safe, border, rare, and outliers. The archetypes are inspired by Napierala and Stefanowski \cite{napierala2016types}. The four categories are determined based on their local neighborhood. We use nearest neighbors to calculate instance similarity. The K-nearest neighbor algorithm (KNN) can be easily implemented with feature embeddings, which are of low dimension and therefore allow the full dataset to be read into working memory - even for large, high resolution datasets such as Places and INaturalist. The KNN algorithm is widely used as a backbone in many imbalanced learning methods \cite{han2005borderline,he2008adasyn,dablain2022deepsmote} and also serves as the basis for many IML techniques \cite{papernot2018deep,lipton2018mythos,keane2019case}. In Section~\ref{sec:arch}, we empirically show that the four archetypes, determined with FE and the KNN algorithm, are closely correlated with model accuracy. 

More broadly, the four archetypes facilitate model, dataset and class complexity understanding. We use $K=5$ to determine the local neighborhood. The "safe" category represents class instances whose nearest neighbors are from the same class ($N_c=4$ or $N_c=5$), where $N_c$ is the number of same class neighbors. Therefore, they are likely homogeneous. The border category are instances that have both same and adversary class nearest neighbors (same class neighbors where $N_c=2$ or $N_c=3$) and likely reside at the class decision boundary. The rare category represents class sub-concepts (same class neighbors where $N_c=1$). Finally, the outlier category are instances that do not have any same class neighbors ($N_c=0$). In the case of the majority class, outliers may indeed represent noisy instances, whereas for the minority class, the model may classify more instances as outliers due to a reduced number of training examples and the model's inability to disentangle their latent representations from adversary classes.

The four archetypes can be used to select instance prototypes that can be visualized and further inspected (see Section~\ref{sec:arch}).

\subsection{Training Set Nearest Adversaries to Gauge Validation Set Error}

In this section, we use the proclivity of neural networks to memorize training data to gauge the classes that the model will have difficulty predicting during inference (i.e., produce false positives) \cite{zhang2021understanding}.

\begin{algorithm}[h]
\scriptsize
\caption{Gauge Validation Set False Positives with Training Set Nearest Adversaries}\label{alg:FP_NA_algo}

\DontPrintSemicolon 
\BlankLine
Extract feature embeddings (FE) for each training set instance.\\
\BlankLine
Run K-Nearest Neighbors on FE.\\
\BlankLine
\For {$each\ instance\ (i)\ in\ Train$}{

Determine the class of each nearest neighbor.\;
Label each neighbor based on class.\;
\If {the\ instance\ is\ a\ False\ Positive}{
Count the number of Adversary Class $(A_c)$ neighbors\;
Store counts\;}
 } 

Standardize the counts by Reference Class $(R_c)$ .\;
Visualize the counts for each $R_c$.\;
\BlankLine
\BlankLine
\vspace{-.2cm}
\end{algorithm}

We believe that the local neighborhood of training instances contains important information about class similarity and overlap. During training, if a CNN embeds two classes in close proximity in latent space, then the model will likely have difficulty disentangling its representations of the two classes during inference \cite{dablain2022understanding,dablain2022efficient}. This failure to properly separate the classes during training will likely lead to false positives at validation and test time.

Based on this insight, we extract feature  embeddings (FE) and their labels from a trained model and use the KNN algorithm to find the K-nearest neighbors of each training instance. If an instance produces a false positive during training, we collect and aggregate the number of nearest adversary class neighbors for each reference class. Then, we visualize the results. See Algorithm~\ref{alg:FP_NA_algo}. 

\begin{table*}[t!]
\vspace{-0.1cm}
\footnotesize
\caption{\textbf{Datasets \& Training}}
\label{tab: data}
\centering
\begin{tabular}{ p{1.4cm}p{.8cm}p{1.2cm}p{1.2cm}
p{.9cm}p{.6cm}p{1.2cm}p{1.2cm}p{.7cm}p{.9cm}}
\toprule

\textbf{Dataset} & \textbf{Number of Classes} &
\textbf{Train \mbox{Examples}} &
\textbf{Test \mbox{Examples}} &
\textbf{Input Dim (pixels)} & \textbf{FE Dim} & \textbf{Imbalance Type} &
\textbf{Max \mbox{Imbalance} Level} &
\textbf{Epochs} &
\textbf{Arch.}\\

\midrule

CIFAR-10 & 10 & 12046 & 10000 & 3072 & 64 & Exponential & 100:1 & 200 & Res-32\\
CIFAR-100 & 100 & 19573 & 5000 & 3072 & 64 & Exponential & 10:1 & 200 & Res-32\\
INaturalist & 13 &  72358 & 14020 & 1440000 & 64 & Natural & 50:1 & 50 & Res-56\\
Places-10 & 10 & 15000 & 5000 & 196608 & 64 & Step & 5:1 & 90 & Res-56\\
Places-100 & 100 & 98072 & 15000 & 196608 & 64 & Exponential & 10:1 & 50 & Res-56\\
\bottomrule

\end{tabular}
\vspace{-.4cm}
\end{table*}

In Section~\ref{sec:nnb}, we show visualizations of this technique and how it correlates with validation set false positives using the Kullback Leibler Divergence. In addition, we compare our method to another method that has been used in imbalanced learning to measure class overlap - Fisher's Discriminant Ratio (FDR):


\begin{equation}FDR=\frac{(\mu_{{FE}_i} - \mu_{{FE}_k})^2}{\sigma_{{FE}_i}^2 + \sigma_{{FE}_k}^2} \label{eq}\end{equation}

In \eqref{eq}, i and k represent pair-wise classes in a dataset and FE is a vector of feature embeddings, where the mean squared difference of latent features (FE) is divided by their variance. As used in \cite{barella2021assessing}, FDR is a measure of how close two classes are, with lower values indicating greater similarity. Thus, like our nearest adversary technique, it can be used to determine class overlap, which can be an indicator of adversary class false positives (i.e., a model will falsely predict an adversary class label if the adversary class FE are in close proximity to the reference class FE).

\subsection{Identify Specific Class Feature Map Overlap}\label{sec:fe_olap}

In the previous section, we examined class overlap at the \textit{instance-level}, by counting the number of adversary class local neighbor instances. Here, we focus on overlapping class \textit{latent features}. Each feature embedding (FE) represents the scalar value of a convolutional feature map (FM) in the last layer of a CNN, after pooling. These FE / FM are naturally indexed and can be extracted in vector form. This natural indexing allows us to identify the FE's with the highest magnitudes across an entire class.

\begin{algorithm}[h!]
\scriptsize
\caption{Identify Specific Class Feature Map Overlap}\label{alg:fe_olap}

\DontPrintSemicolon 
\BlankLine
Extract feature embeddings (FE) for each training set instance.\\
\BlankLine

\For {$each\ class\ (c)\ in\ Train$}{
Calculate the mean of each feature embedding\;
Select the top K FE (e.g., K=10) based on mean\;
 } 

Visualize.\;
\BlankLine
\BlankLine
\vspace{-.2cm}
\end{algorithm}

Each FE has a magnitude. For each class, the FE magnitudes can be aggregated and averaged. Then, the FE with the largest magnitudes can be selected (the top-K FE). If two classes place a high magnitude on a FE / FM with the same index position, then this FE is important for both classes. See Algorithm~\ref{alg:fe_olap}.  

We should remember that a CNN's final classification layer, when trained with cross-entropy loss, or one of its variants, casts its decision based on linear classification. The final, fully connected layer makes a prediction based on the class with the largest logit (highest magnitude). This logit, in turn, is the sum of FE / FM magnitudes multiplied by class weights. Therefore, the individual FE / FM magnitudes are important to class prediction.  Recent work shows that the FE magnitude is relatively more important than the final layer classification weights in forming a class prediction \cite{dablain2022understanding}. We use this insight to identify the top-K FE for each class, and to determine where the FE / FM indexes overlap, which may lead to false positives. 

In Section~\ref{sec:olap}, we provide visualizations of this method, along with suggestions for how it can be used by model users and developers.

\begin{figure*}[!hb]
   \vspace{-0.3cm}
  \centering
 
  \subfloat[CIFAR- 10 (exp), 100:1]{\includegraphics[width=0.2\textwidth]{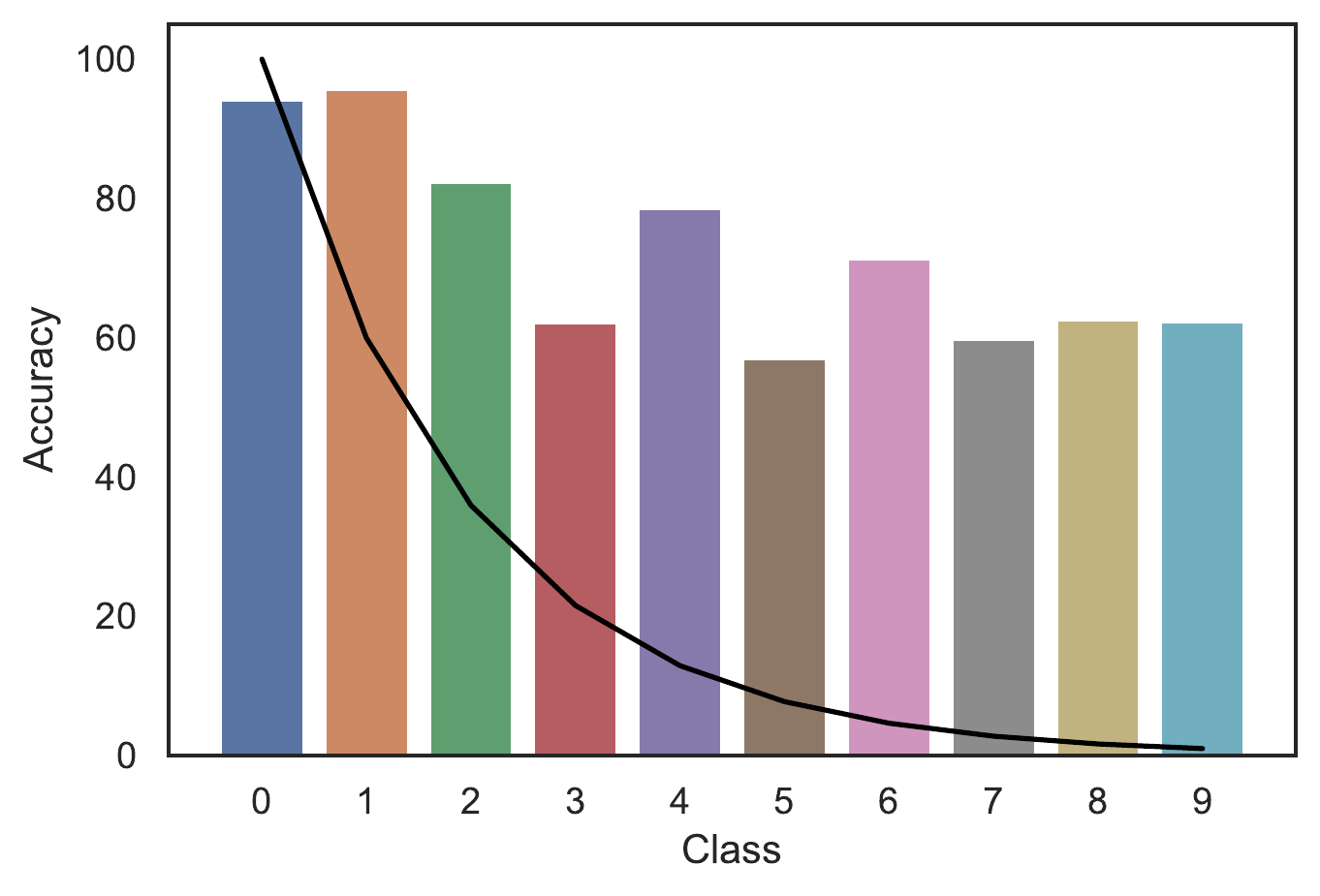}\label{fig:f2}}
   \hfill
  \subfloat[Places (step), 5:1]{\includegraphics[width=0.2\textwidth]{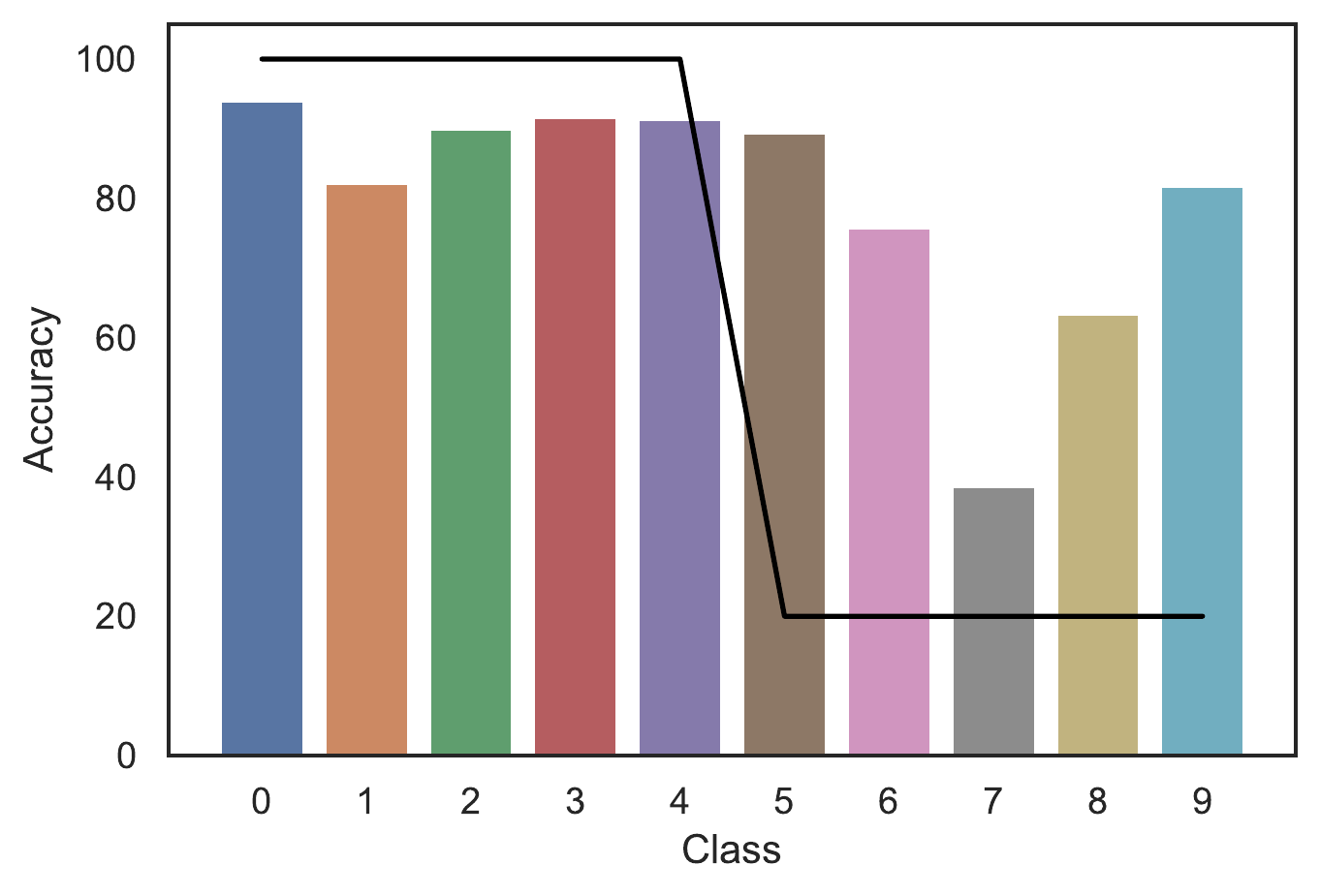}\label{fig:f3}}
   \hfill
  \subfloat[INaturalist (natural), 50:1]{\includegraphics[width=0.2\textwidth]{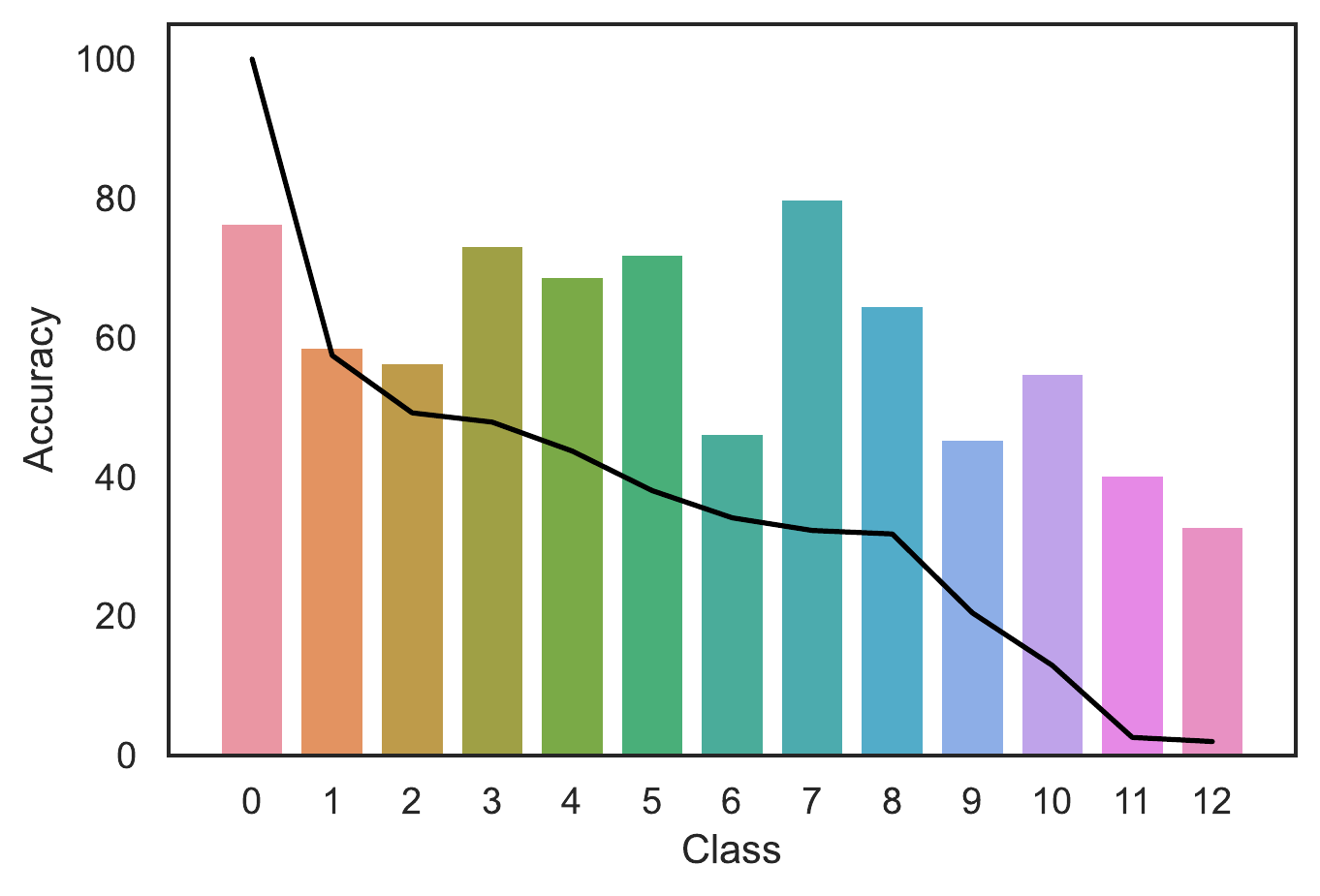}\label{fig:f4}}
  \hfill
  \subfloat[CIFAR-100 (exp, 10:1)]{\includegraphics[width=0.2\textwidth]{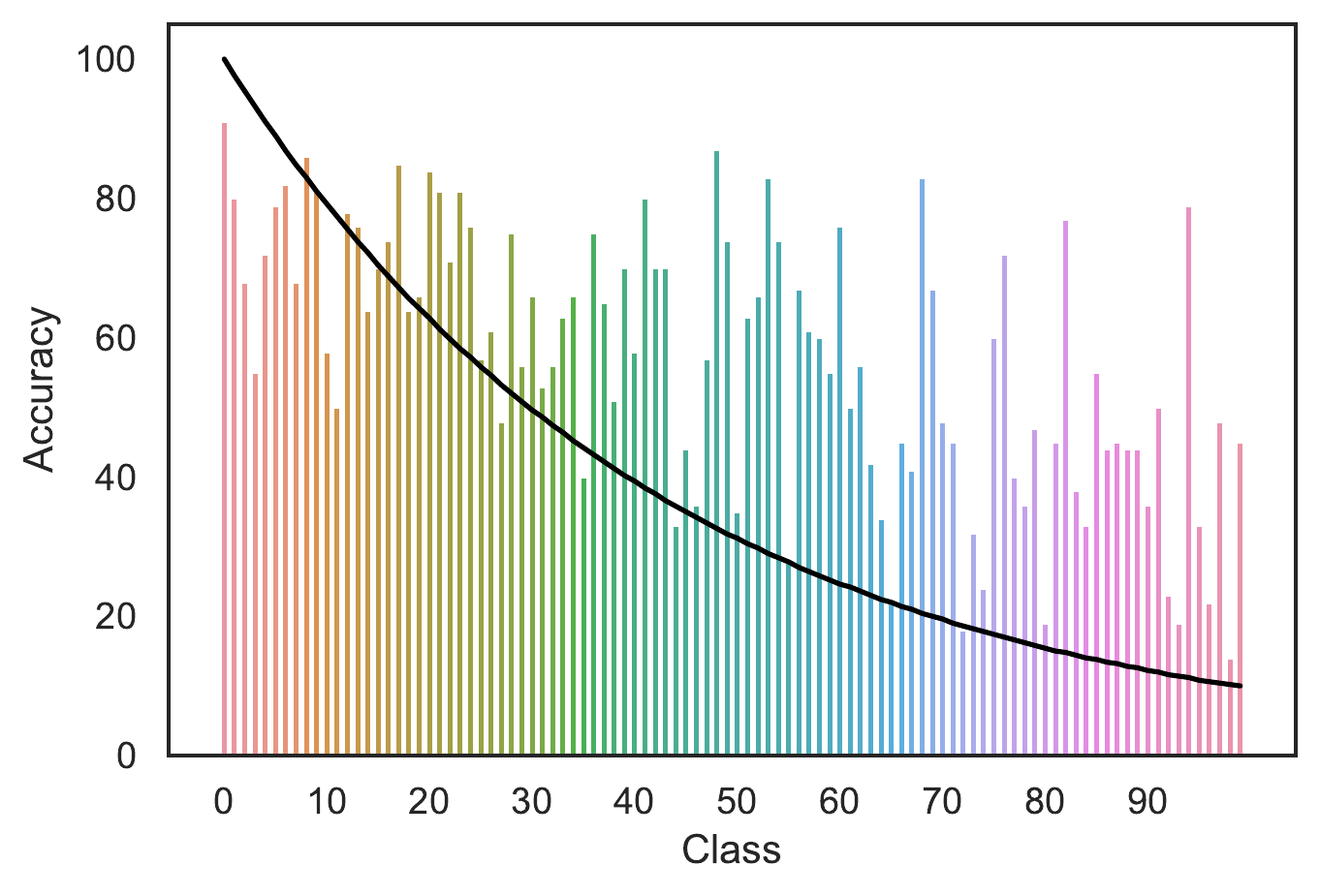}\label{fig:f1}}
  \hfill
  \subfloat[Places-100 (exp, 10:1)]{\includegraphics[width=0.2\textwidth]{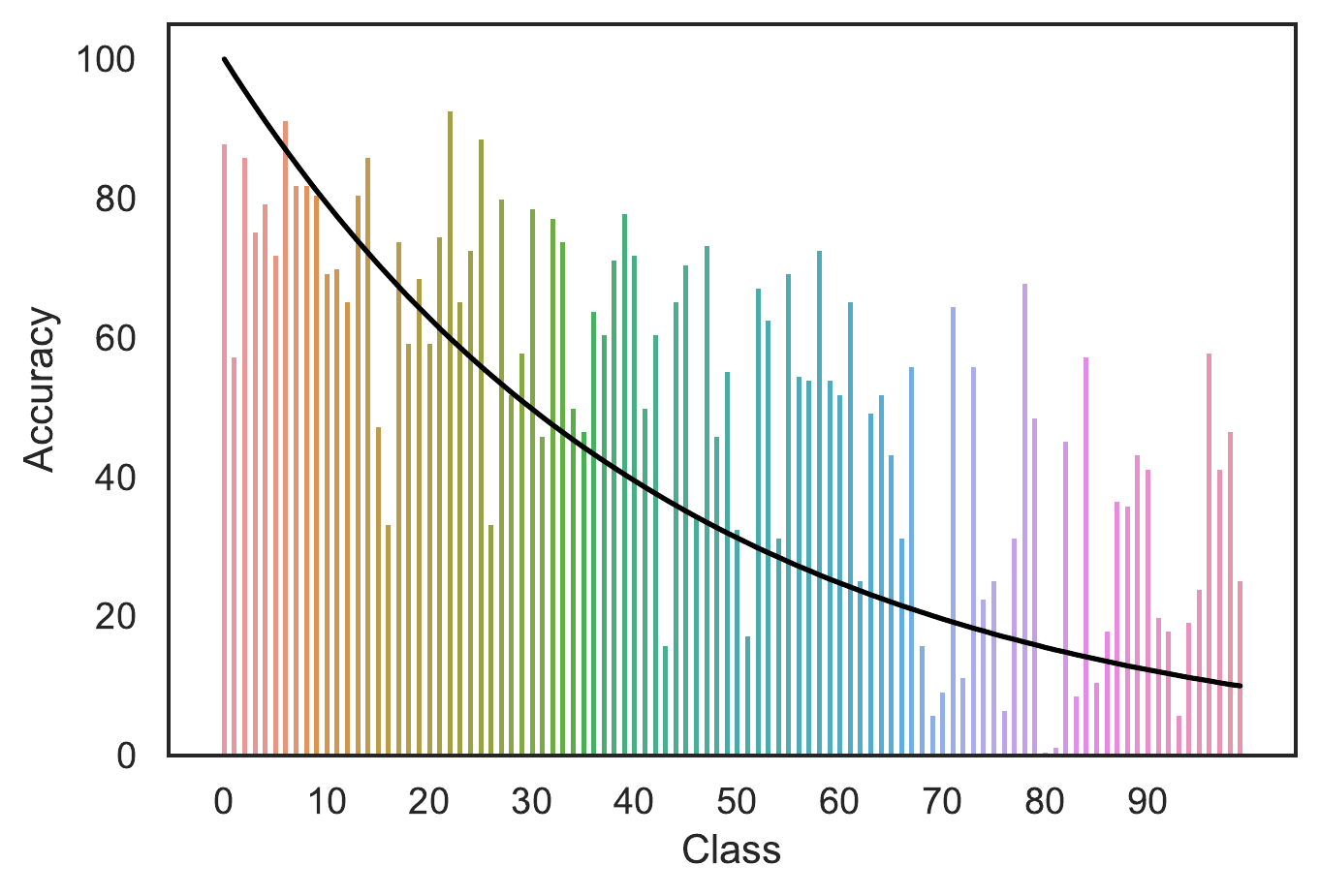}\label{fig:f1a}}
  \caption{This figure presents individual class accuracies for 5 datasets, based on exponential (exp), step and natural imbalance levels, with the maximum imbalance level listed in each subcaption. This illustration serves as a high-level indicator of dataset complexity that can be used for preliminary data inspection purposes. For all datasets, although accuracy is "directionally" impacted by imbalance (depicted with black lines), it does \textit{not} monotonically follow class imbalance, which indicates that other factors, such as class overlap, noise and sub-concepts, may be in play.}
  \label{fig_accur}
  \vspace{-0.3cm}
\end{figure*}

\subsection{Feature Density}

As discussed in Section~\ref{sec:RW}, a number of resampling methods seek to balance the density of class examples along the decision boundary. These methods are typically applied to balance the number of class \textit{instances}.

\begin{algorithm}[h!]
\scriptsize
\caption{Feature Density}\label{alg:density}

\DontPrintSemicolon 
\BlankLine
Extract feature embeddings (FE) for each training set instance.\\
\BlankLine
Select a reference class $(R_C)$.\\
\BlankLine
Identify the Top-K FE for $R_C$  $(Top_{RC})$\\
\BlankLine
\For {each\ Adversary\ Class\ $(A_C)$ }{
\For {each\ instance\ in\ $A_C$ }{
Count the number of instances where $Top_{AC} \in Top_{RC}$\;}}
Divide each count by the number of instances in {$R_C$}.\;
Visualize.\;
\BlankLine
\BlankLine
\vspace{-.2cm}
\end{algorithm}

Given the importance of the relative number of majority and minority class instances to model accuracy, we extend and refine the concept of class density to class \textit{latent features}. For each reference class in a training set, we track the number of adversary class instances where the top-K FE indices of the adversary class match the top-K FE indices of the reference class (we use $K=10$). We base the top-K FE on magnitude since CNN classifiers vote for a class via the magnitude of the largest logit. This provides us with a view of the most important reference class feature maps (as represented by the FE scalars) that overlap with an adversary class. This also provides insight into not only the specific latent features that the model entangles; but, as discussed in Section~\ref{sec:den}, it also furnishes opportunities for resampling methods based on specific features, instead of instances. Thus, it introduces the possibility of over- or under-sampling to balance class densities based on latent features, instead of instances.

\subsection{Colors that Define Classes}\label{sec:texture}

Existing IML methods that trace CNN decisions to pixel space via gradient techniques track salient pixel \textit{locations} for \textit{single} image instances. They display a virtual black and white source image (black to indicate high pixel saliency to a CNN's prediction and white to indicate low saliency). We make use of a gradient saliency technique commonly used in IML, which was developed by Simonyan et al. \cite{simonyan2013deep}. However, we modify it to trace a prediction to a pixel location only so that we can extract the RGB pixel values at that location.

\begin{algorithm}[h!]
\scriptsize
\caption{Colors that Define Classes}\label{alg:texture}

\DontPrintSemicolon 
\BlankLine
Select a Reference Class $(R_C)$.\\
\BlankLine

\For {$each\ instance\ (i)\ in\ R_C$}{

Identify Top-K input pixels by back propagating the gradient\;
Group pixels into bins based on color bands\;
Adaptively update bin means\;
 } 

Visualize.\;
\BlankLine
\BlankLine
\vspace{-.3cm}
\end{algorithm}

We use RGB pixel values, instead of locations, in our visualizations because, on a class basis, over many images, the locations themselves are likely to be meaningless due to changes in pose, translation, rotation, etc. We collect the top-K RGB pixel values for each instance in a training set and also the instance labels. For purposes of our illustrations in Section~\ref{sec:text_exp}, we select the top 10\% most salient pixels.  We partition the collected pixels into bins based on the color spectrum (e.g., black, orange, red, green, blue, light blue, white, etc.). See Algorithm~\ref{alg:texture}.

Storing such information for large image datasets may be difficult to maintain in working memory. For example, the dimensions of the tracked RGB pixels could be $N_I$ X 3 X $N_P$, where $N_I$ is the number of images in the training set, 3 is the number of channels, and $N_P$ is the number of top-K pixels per instance. Therefore, we collect the top-K pixels for each instance and adaptively update a single mean for each color bin. This approach transforms the working memory storage requirements to 14 X 3, where the number of color bins is 14 for 3 color channels. We use the bins to track counts and develop a histogram for visual display. We only use the color means for purposes of displaying the central color of each bin. See Section~\ref{sec:text_exp} for visualizations.

\begin{figure*}[t!]
   \vspace{-.5cm}
  \centering
  \subfloat[CIFAR-10]{\includegraphics[width=0.2\textwidth]{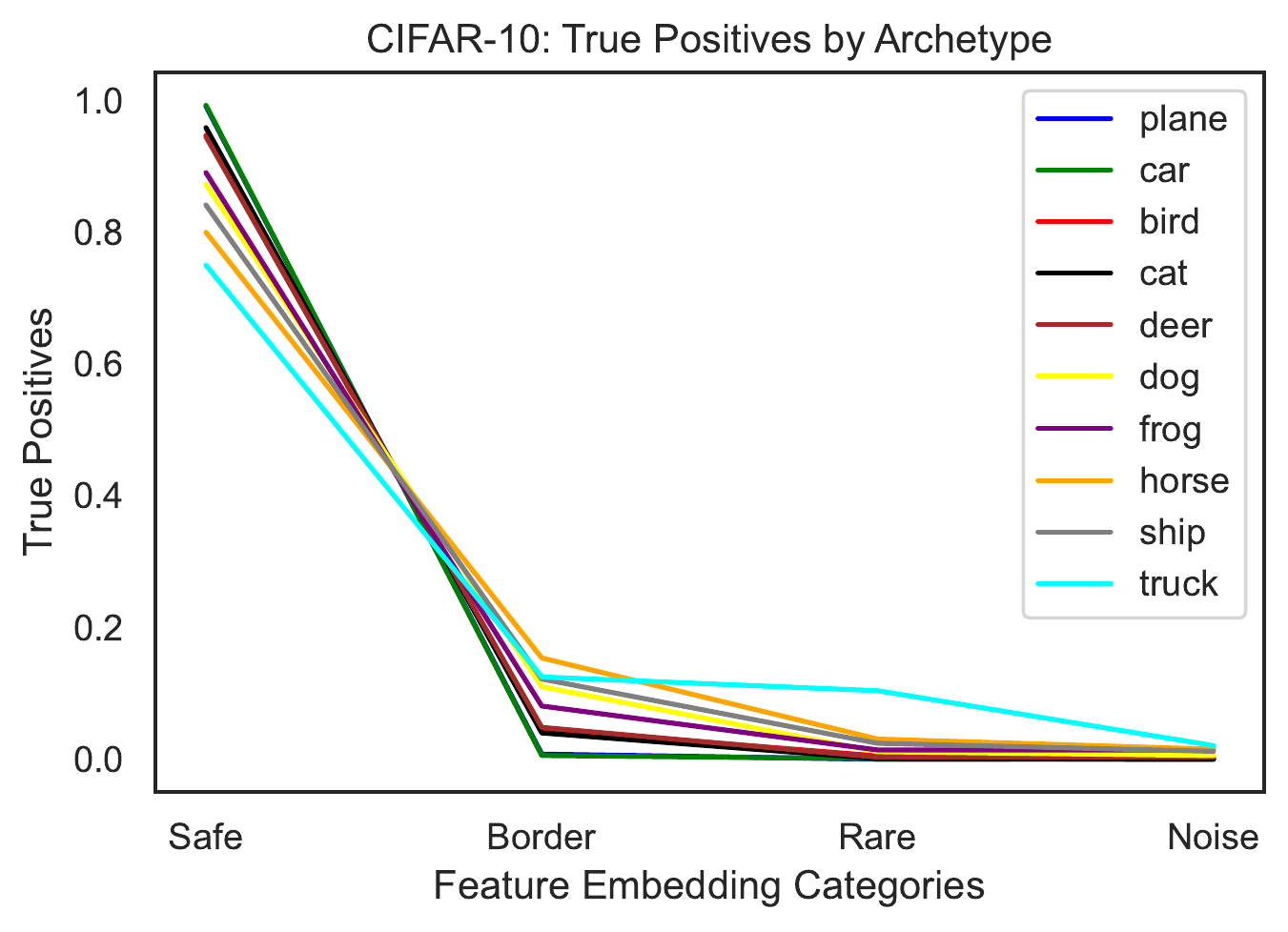}\label{fig:f5}}
  \hfill
  \subfloat[CIFAR-100]{\includegraphics[width=0.2\textwidth]{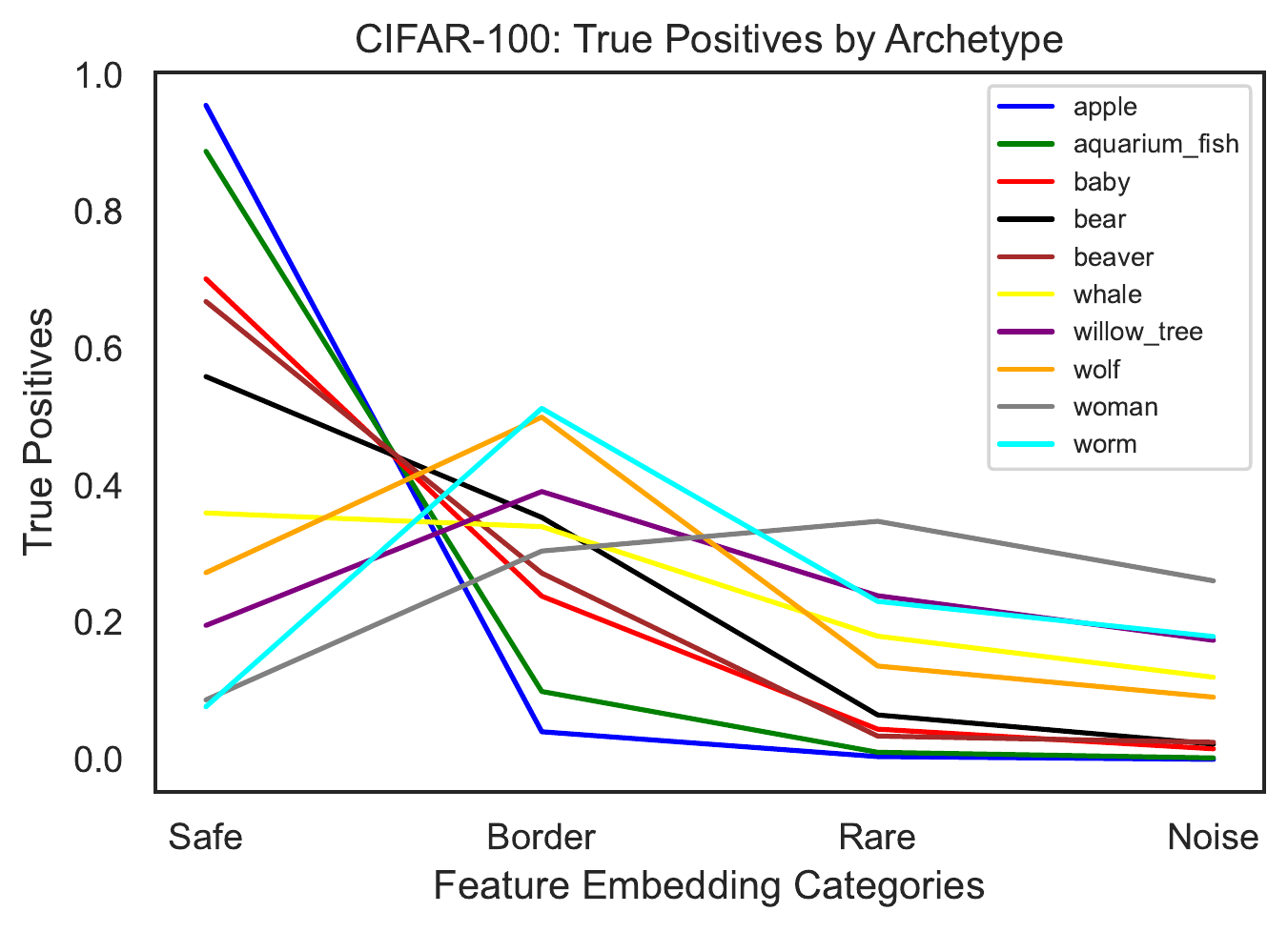}\label{fig:f6}}
   \hfill
  \subfloat[Places-10 (step), 5:1]{\includegraphics[width=0.2\textwidth]{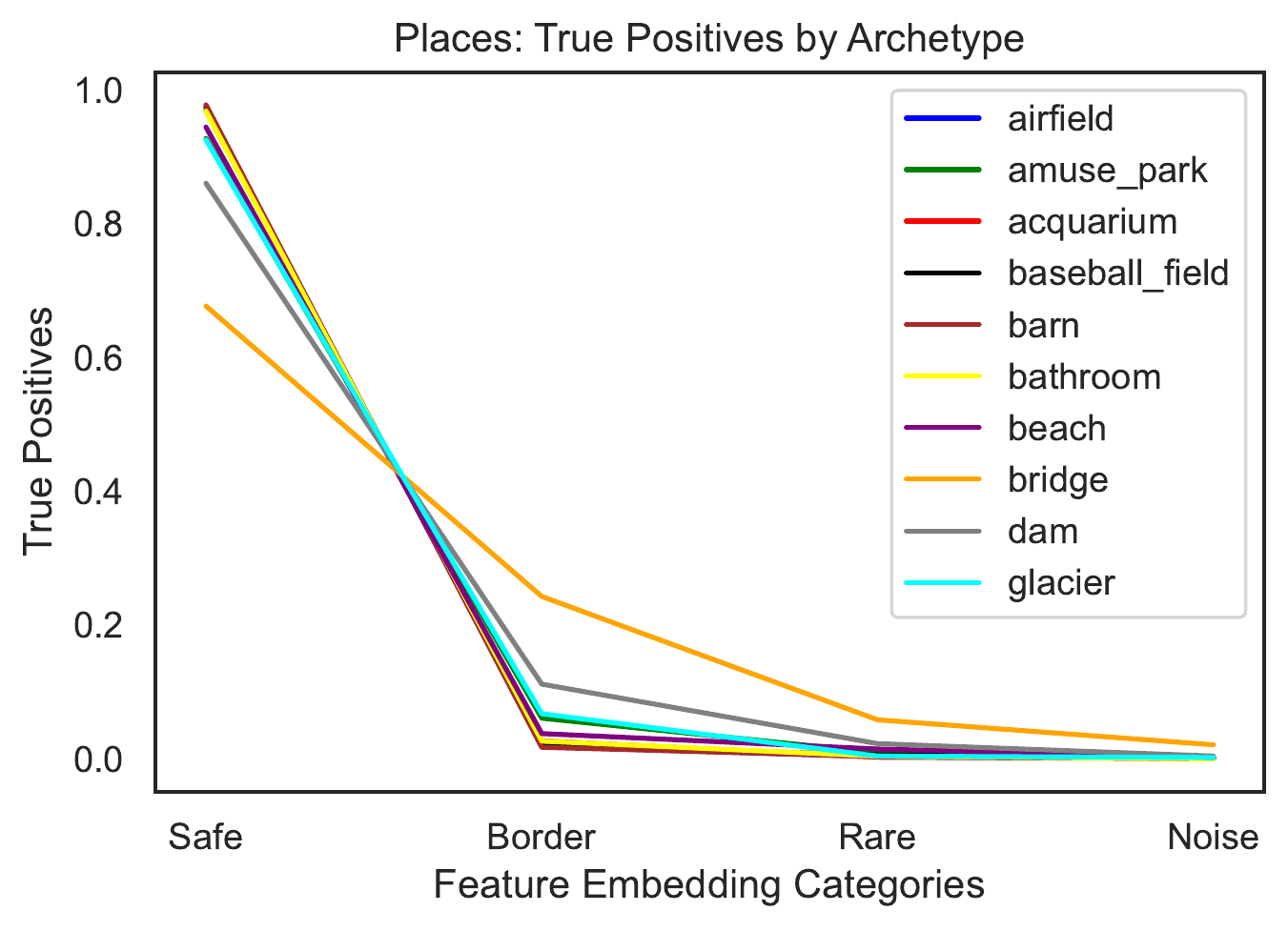}\label{fig:f7}}
   \hfill
  \subfloat[INaturalist]{\includegraphics[width=0.2\textwidth]{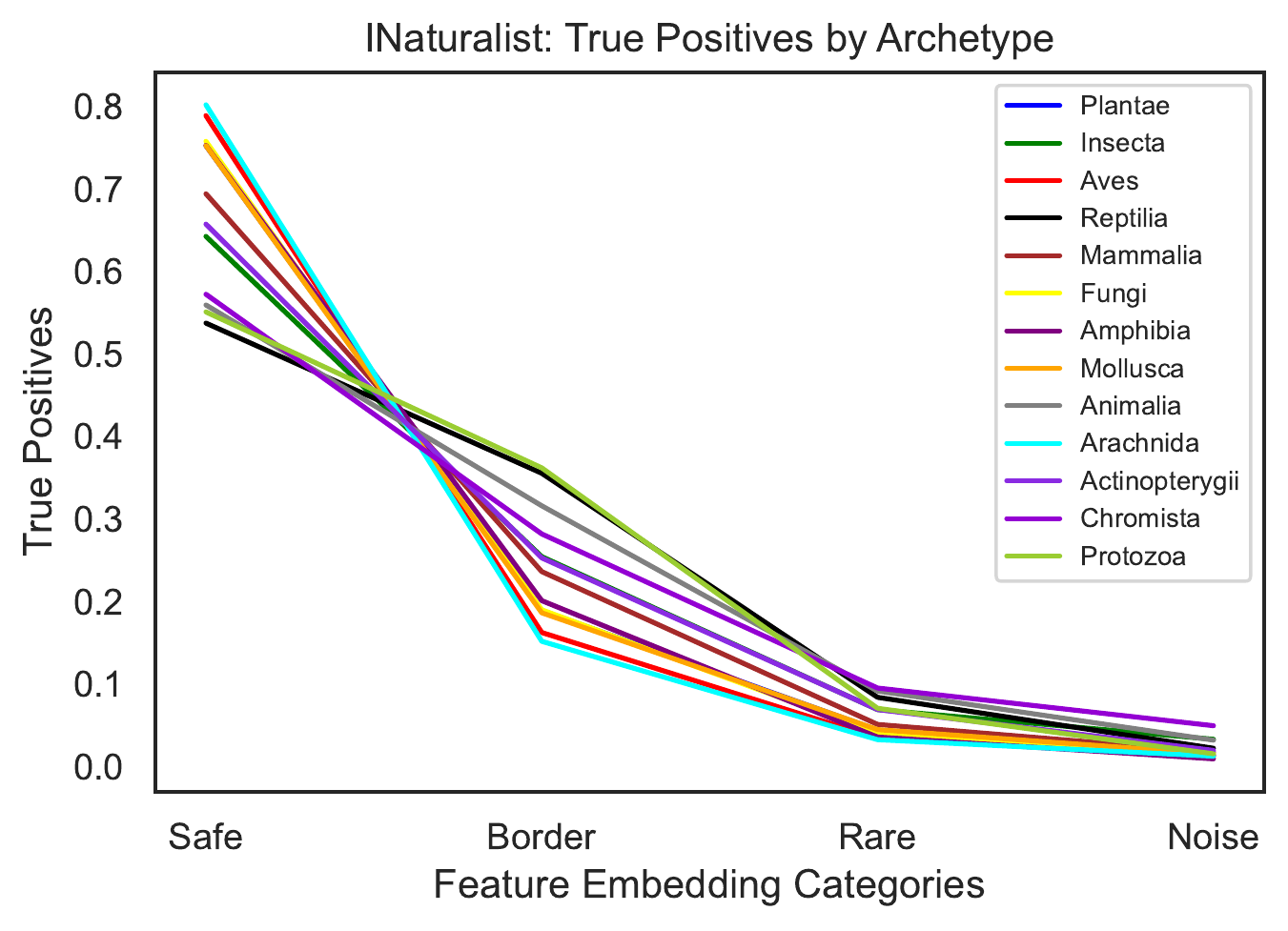}\label{fig:f8}}
 \hfill
  \subfloat[Places-100]{\includegraphics[width=0.2\textwidth]{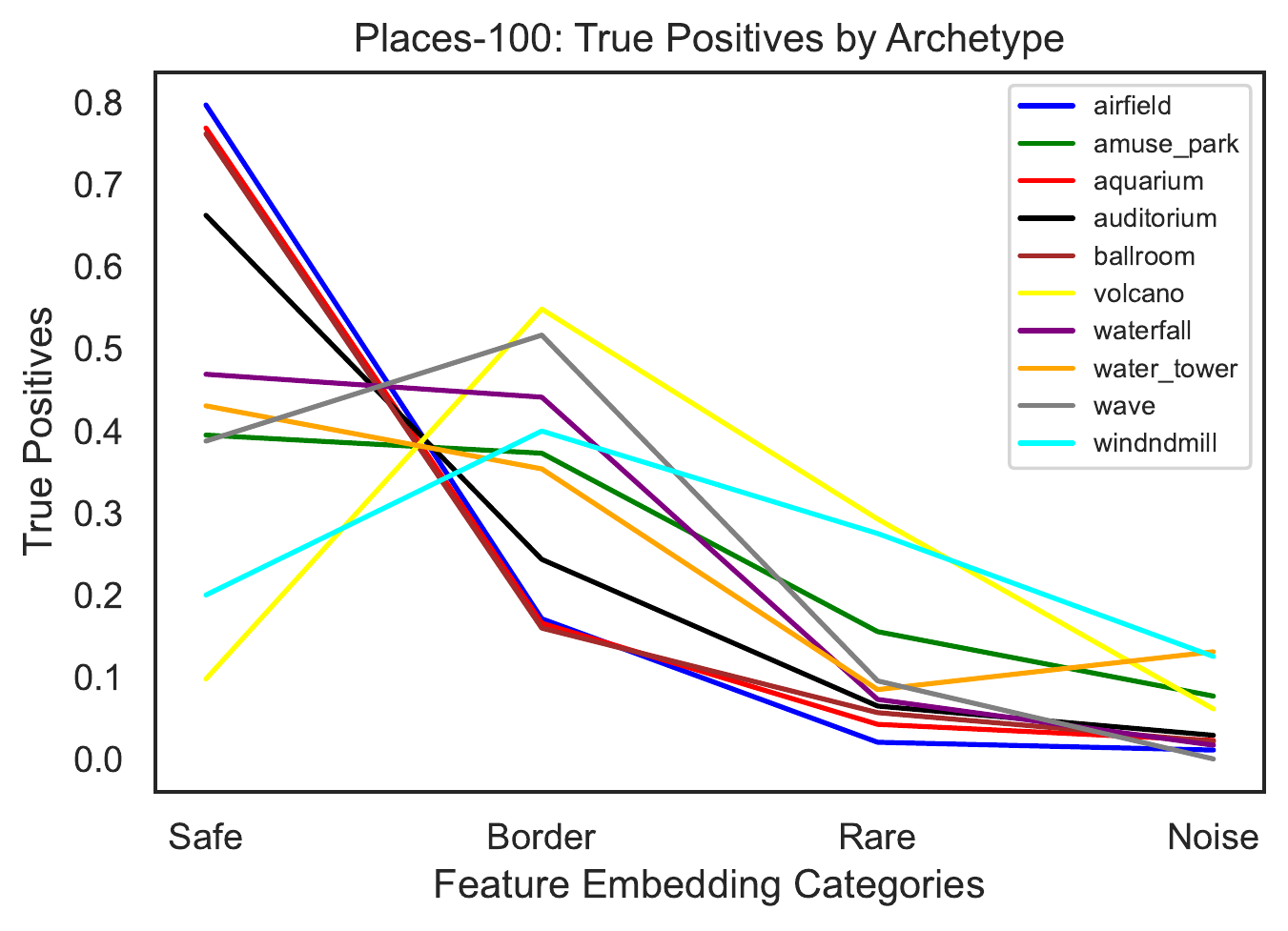}} 
  \caption{This figure shows the percentage of True Positives (TPs) of the 4 archetypes: safe, border, rare and outliers in each training set. For CIFAR-100 and Places-100, the classes with the 5 largest number of examples and the 5 fewest are shown to make the visualization interpretable. With the exception of the extreme minority classes in CIFAR-100 and Places-100, the safe category contains the highest percentage of TPs, with a gradual decline over the other archetypes.}
  \label{fig_arch}
  \vspace{-0.4cm}
\end{figure*}

\begin{figure}[!b]
   \vspace{-0.6cm}
  \centering
  \subfloat[Safe]{\includegraphics[width=0.12\textwidth]{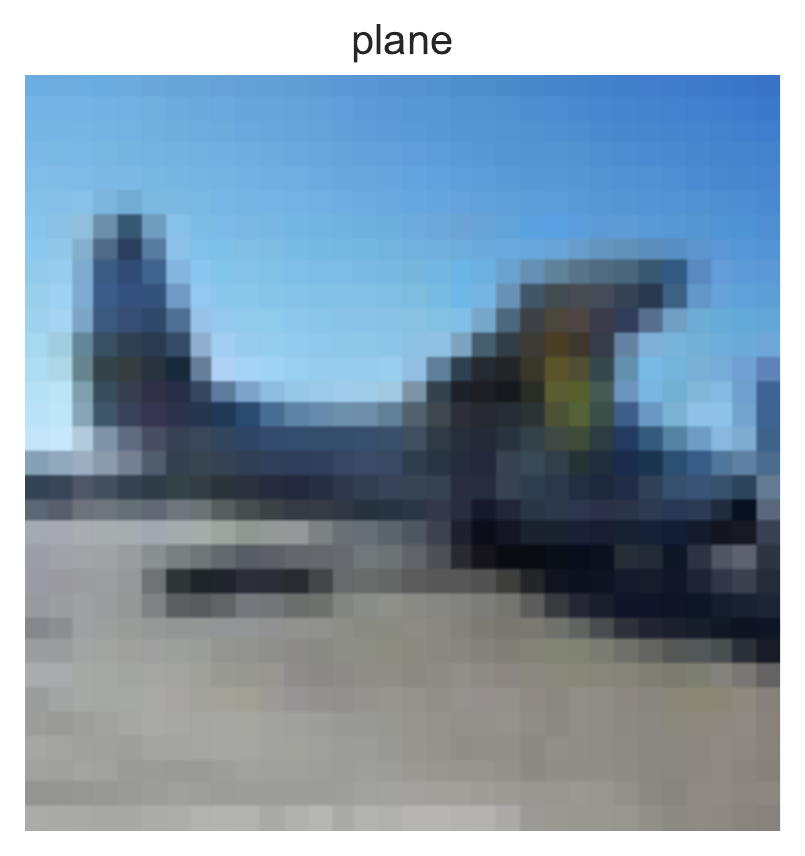}\label{fig:f9}}
  \hfill
  \subfloat[Border]{\includegraphics[width=0.12\textwidth]{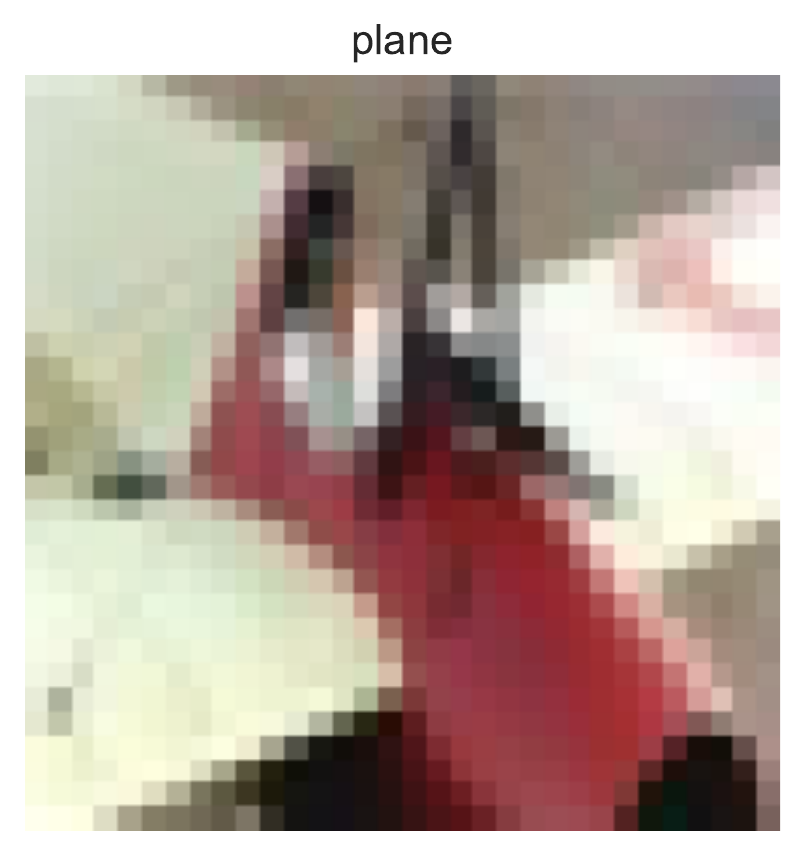}\label{fig:f10}}
   \hfill
  \subfloat[Rare]{\includegraphics[width=0.12\textwidth]{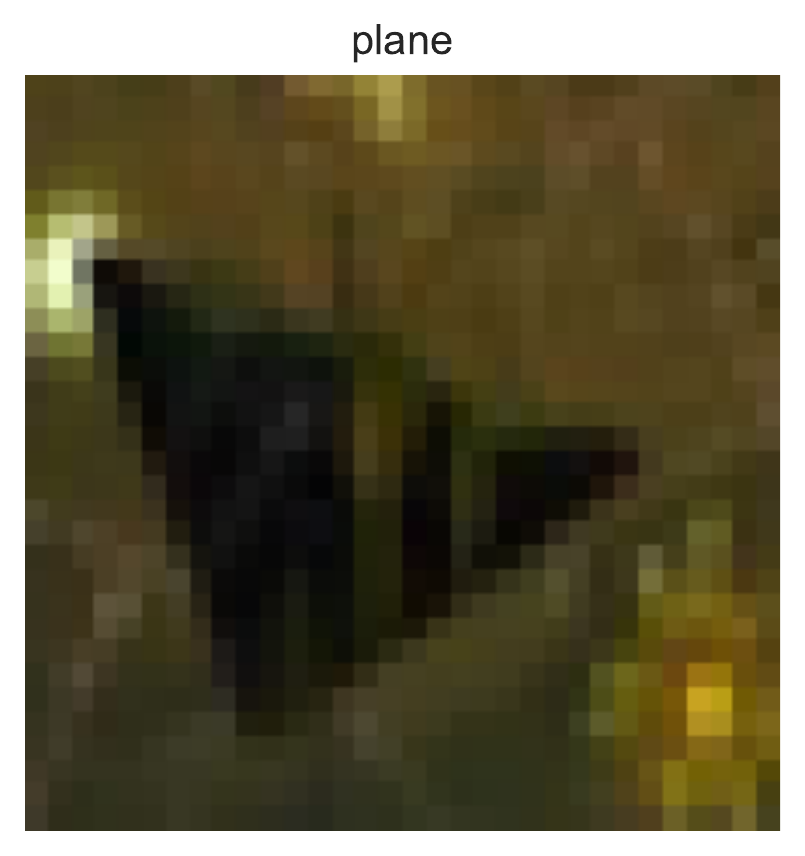}\label{fig:f11}}
   \hfill
  \subfloat[Outlier]{\includegraphics[width=0.12\textwidth]{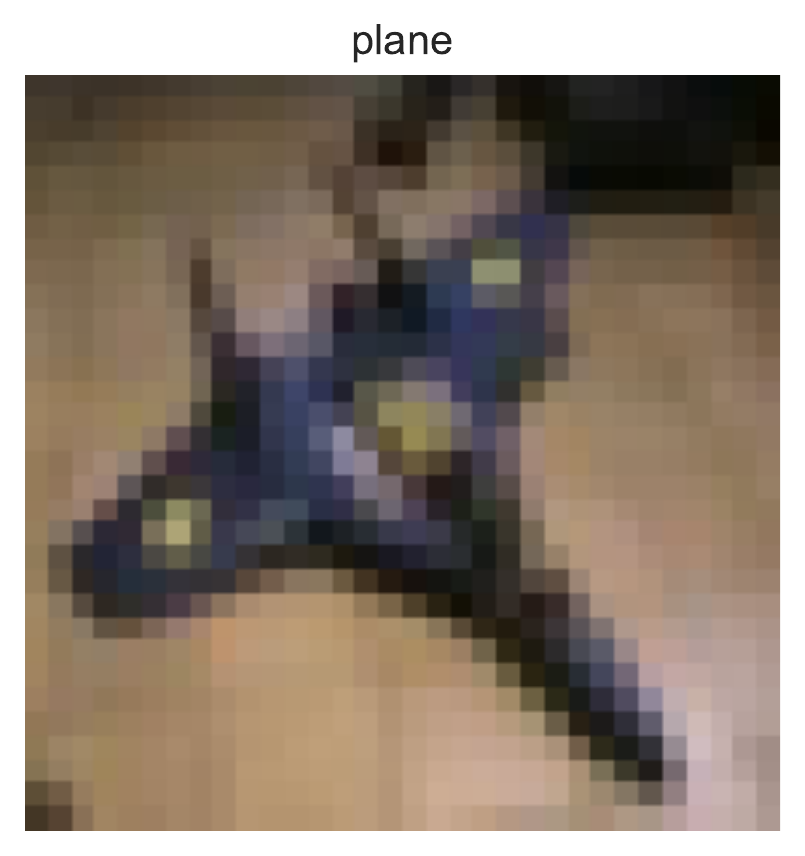}\label{fig:f12}}
  \hfill
  \subfloat[Safe]{\includegraphics[width=0.12\textwidth]{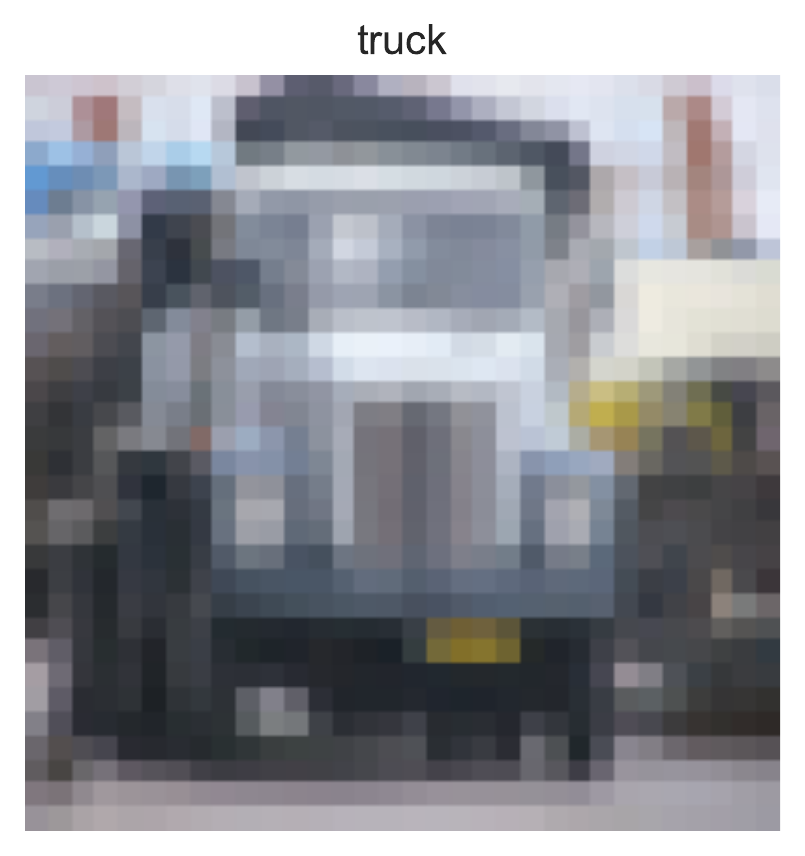}\label{fig:f13}}
  \hfill
  \subfloat[Border]{\includegraphics[width=0.12\textwidth]{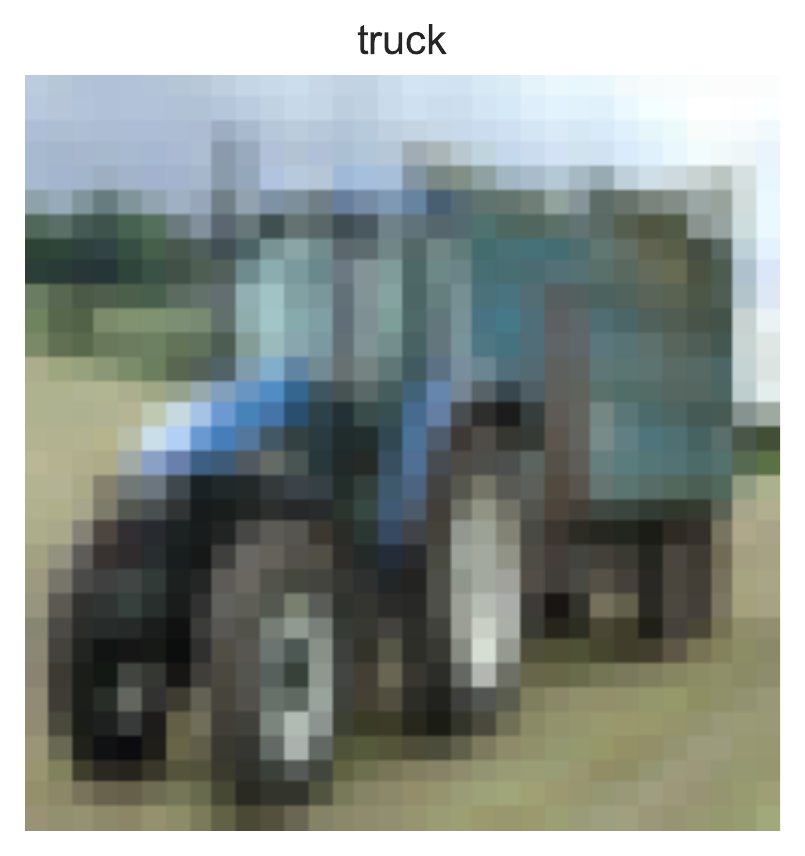}\label{fig:f14}}
   \hfill
  \subfloat[Rare]{\includegraphics[width=0.12\textwidth]{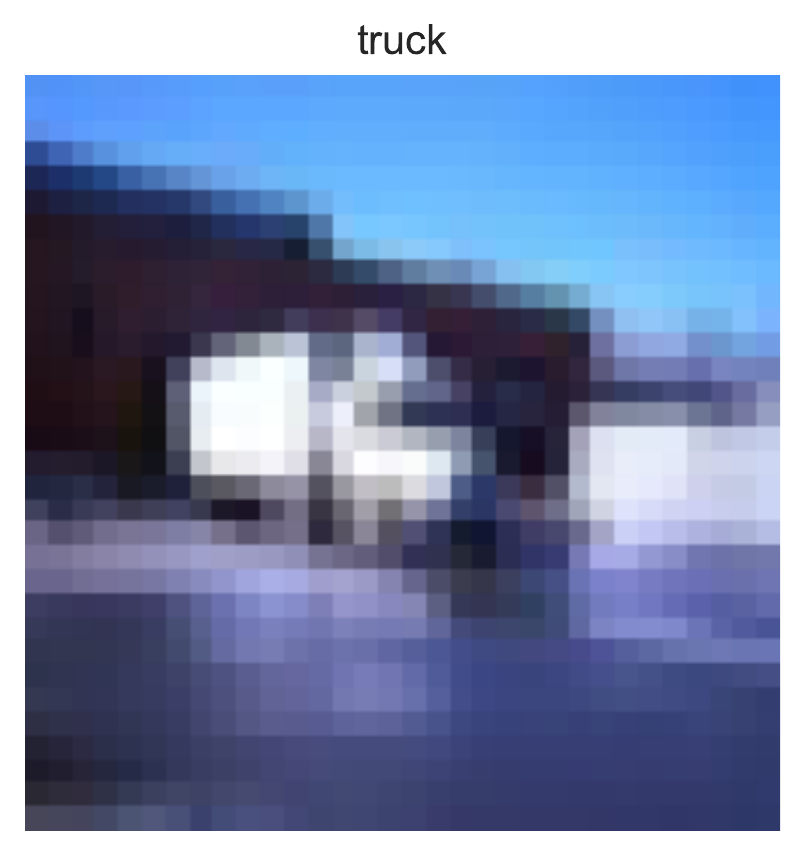}\label{fig:f15}}
   \hfill
  \subfloat[Outlier]{\includegraphics[width=0.12\textwidth]{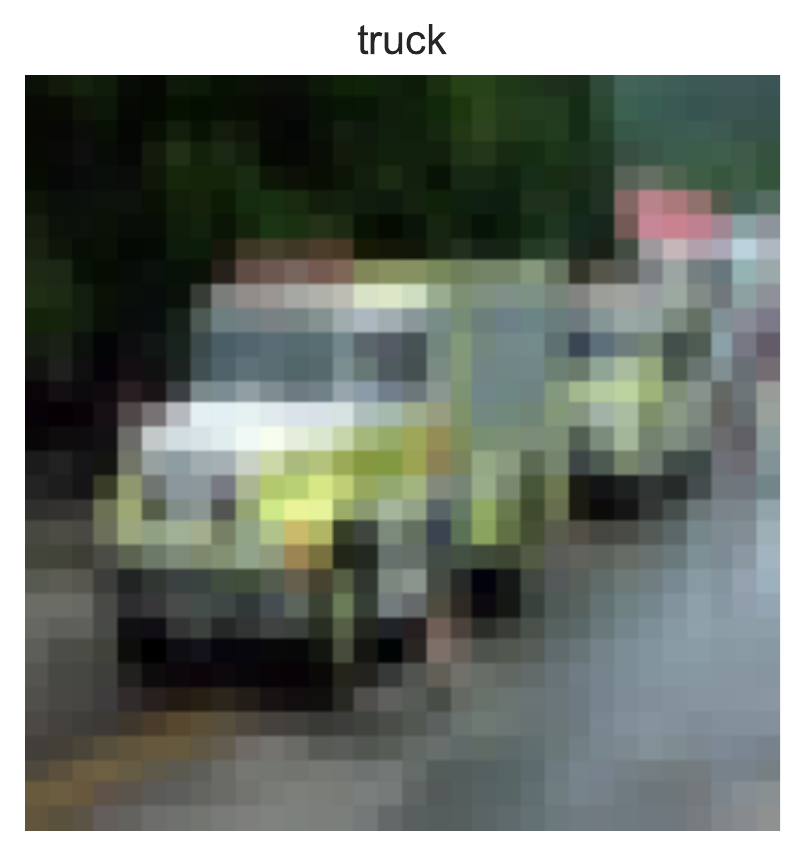}\label{fig:f16}}
 
  \caption{This figure displays the safe, border, rare and outlier prototypes for 2 classes in the CIFAR-10 dataset.  Figures (a) - (d) are from the class with the largest number of examples (airplanes) and (e) - (h) are from the class with the fewest number of examples (trucks). These figures allow model users and imbalanced learning algorithm developers to visualize representative examples of class archetypes. In the case of the majority class, outliers may indeed represent noisy instances, whereas for the minority class, the model may classify more instances as outliers due to a reduced number of training examples. In addition, visualizing representative class images may help flag hidden CNN classifier biases. For example, for a human, who preferences shape in object recognition, the difference between the plane safe and outlier categories may not be immediately apparent. See the CNN color bias explanation - Section~\ref{sec:text_exp} and Figure~\ref{fig_6_fe_texture} - for possible answers (hint: there's no blue sky in the outlier).}
  \label{fig_arch_viz}
  \vspace{-0.4cm}
\end{figure}

\section{Experiments \& Results}
\label{sec:exper}

\subsection{Experimental Set-up}

To illustrate the application of our framework, we select five image datasets: CIFAR-10, CIFAR-100 \cite{krizhevsky2009learning}, Places-10, Places-100 \cite{zhou2017places} and INaturalist \cite{van2018inaturalist}. For each dataset, we use different types and levels of imbalance to highlight varied applications of our framework (see Table~\ref{tab: data} for dataset details).  To make training tractable, we limit Places to 10 and 100 classes and INaturalist to its 13 super-categories.  CIFAR-10 and CIFAR-100 are trained with a Resnet-32 \cite{he2016deep} backbone and Places and INaturalist with a Resnet-56. Although a Resnet architecture is used for our experiments, any CNN architecture that imposes dimensionality reduction should work (e.g., a Densenet \cite{huang2017densely} likely would not facilitate the use of lower dimensional feature embeddings). We adopt a training regime employed by Cao et al. \cite{Cao:2019}. Except where noted, all models are trained on cross-entropy loss with a single NVIDIA 3070 GPU. As discussed in the following sections, in several cases, we train models with a cost-sensitive method (LDAM) \cite{Cao:2019} to show how visualizations of both baseline and cost-sensitive algorithms can be used for comparative purposes to assess specific areas of improvement or under-performance.

Next, we illustrate each element of our framework, using various imbalanced image datasets and algorithms, where appropriate.

\subsection{Individual Class Accuracy} \label{sec:acc}

As a high-level visualization, we first show
 individual class accuracy for 5 datasets, along with the class imbalance level (black lines). See Figure~\ref{fig_accur}. These figures provide a more detailed view of class accuracy than the single dataset metrics commonly used in imbalanced learning, such as balanced accuracy, macro F1 measure, or geometric mean. When displayed along with the class imbalance ratio level (black lines), it also provides clues as to data complexity. For example, in Figure~\ref{fig_accur}, for all datasets, class accuracy does not strictly follow imbalance. Although there is a general decline along the downward sloping imbalance level, the relationship is not exact, which implies that other factors, such as class overlap, sub-concepts and outliers may be in play. 
 
 \textbf{Use Cases.} This chart can be applied after training and before the validation or test phases, by imbalanced learning algorithm developers. It can be used  to determine how their algorithm affects specific classes compared to baselines. It can also be used by model users during the test phase to visualize how a model performs on specific classes of interest.
 
 In subsequent sections, we will take a closer look at how other factors besides class imbalance, such as instance and latent feature overlap, affects accuracy and suggest potential ways to use visual class overlap explanations as guides to model improvement.

\subsection{Class Archetypes}\label{sec:arch}
Figure~\ref{fig_arch} shows the percentage of true positives (TP) for each class in 5 training datasets. The TPs are grouped based on class archetypes: safe, border, rare and outliers. For 3 of the datasets (CIFAR-10, Places-10, and INaturalist), the safe category contains the greatest percentage of TPs relative to the total number of instances in the group. For the other 2 datasets (CIFAR-100 and Places-100), only the 10 classes with the most and least number of training examples are shown (5 each) to make the visualization less busy. For these two datasets, the majority classes generally exhibit the same relationships for the safe category as noted above; however, the minority classes sometimes show greater relative TPs for the border than the safe category. We hypothesize that the reason for this discrepancy is that the model has not learned to disentangle class features for minority classes due to fewer instances and the larger number of classes. Thus, the model is not able to develop latent features for a clearly homogeneous safe class; rather, it memorizes individual training instances which tend to be more similar to adversary classes (i.e., greater number of adversary neighbors) than members of their own class. 

In Figure~\ref{fig_arch_viz}, we select a prototypical instance from the safe, border, rare and outlier categories for the majority class (airplanes) and the minority class (trucks) from CIFAR-10 for visualization purposes. In large image datasets, it may not be obvious which examples are representative of the overall class (safe examples), which examples reside on the decision boundary (border), and which instances may be sub-concepts or outliers. For each of these categories and classes, we select the most central prototype, using the K-medoid algorithm, and visualize them. 

These visualizations can help identify potential issues that require further investigation of specific classes. For example, in the case of airplanes, it may not be immediately apparent to a human, who preferences shape over texture, why the outlier example is different from the safe prototype in Figure~\ref{fig_arch_viz}. This seeming incongruence can serve as a flag for model users and algorithm developers. As discussed in more detail in Section~\ref{sec:text_exp}, we conjecture that this seeming incongruity is due to a CNN's preference of texture over shape when distinguishing classes (i.e., there is no blue sky in the outlier airplane prototype).

\begin{figure*}[t!]
   \vspace{-0.2cm}
  \centering
  \subfloat[INaturalist: Validation False Positives by Class]{\includegraphics[width=0.49\textwidth]{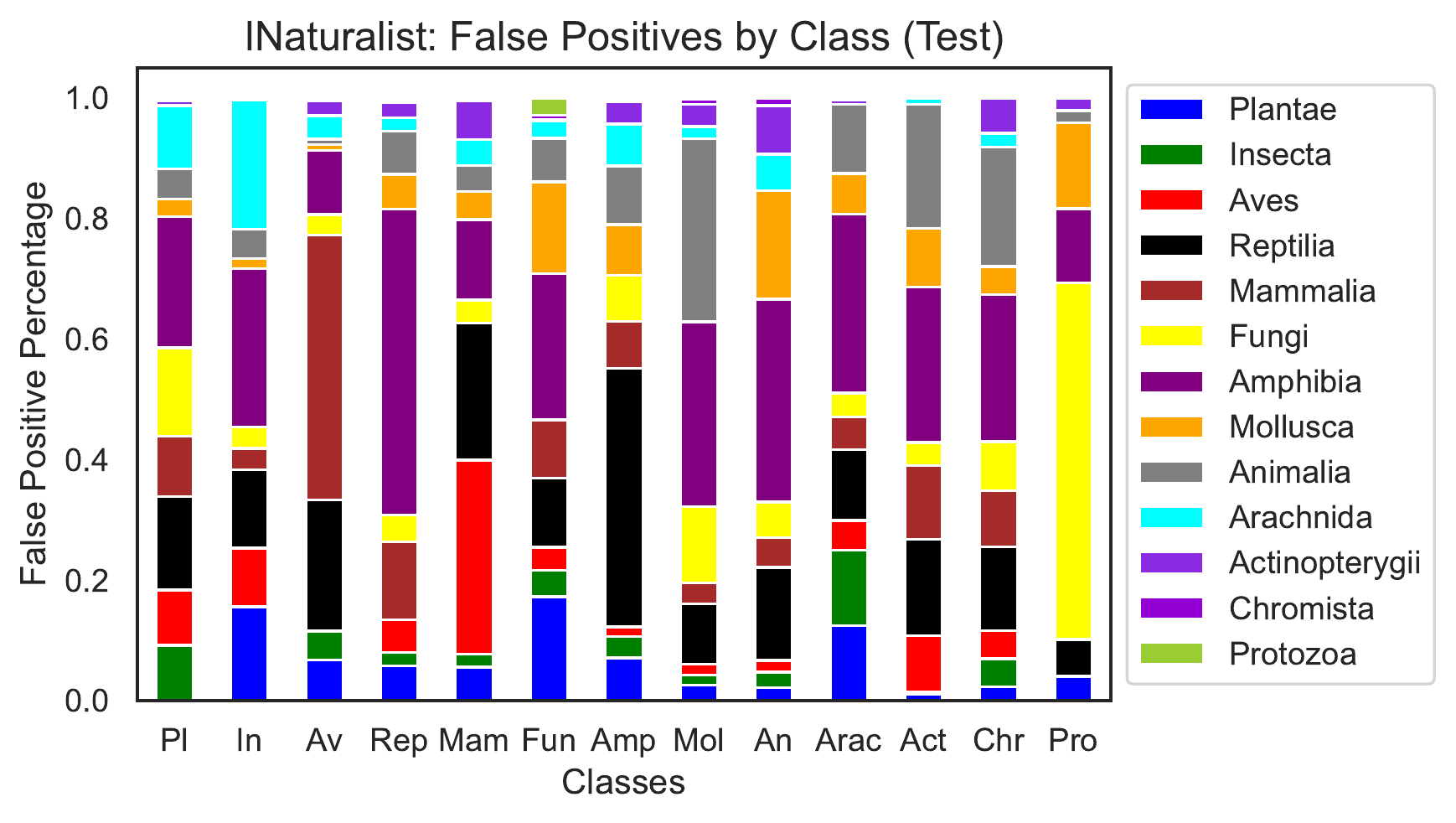}\label{fig:f17}}
  \hfill
  \subfloat[INaturalist: Training Nearest Adversaries by Class]{\includegraphics[width=0.49\textwidth]{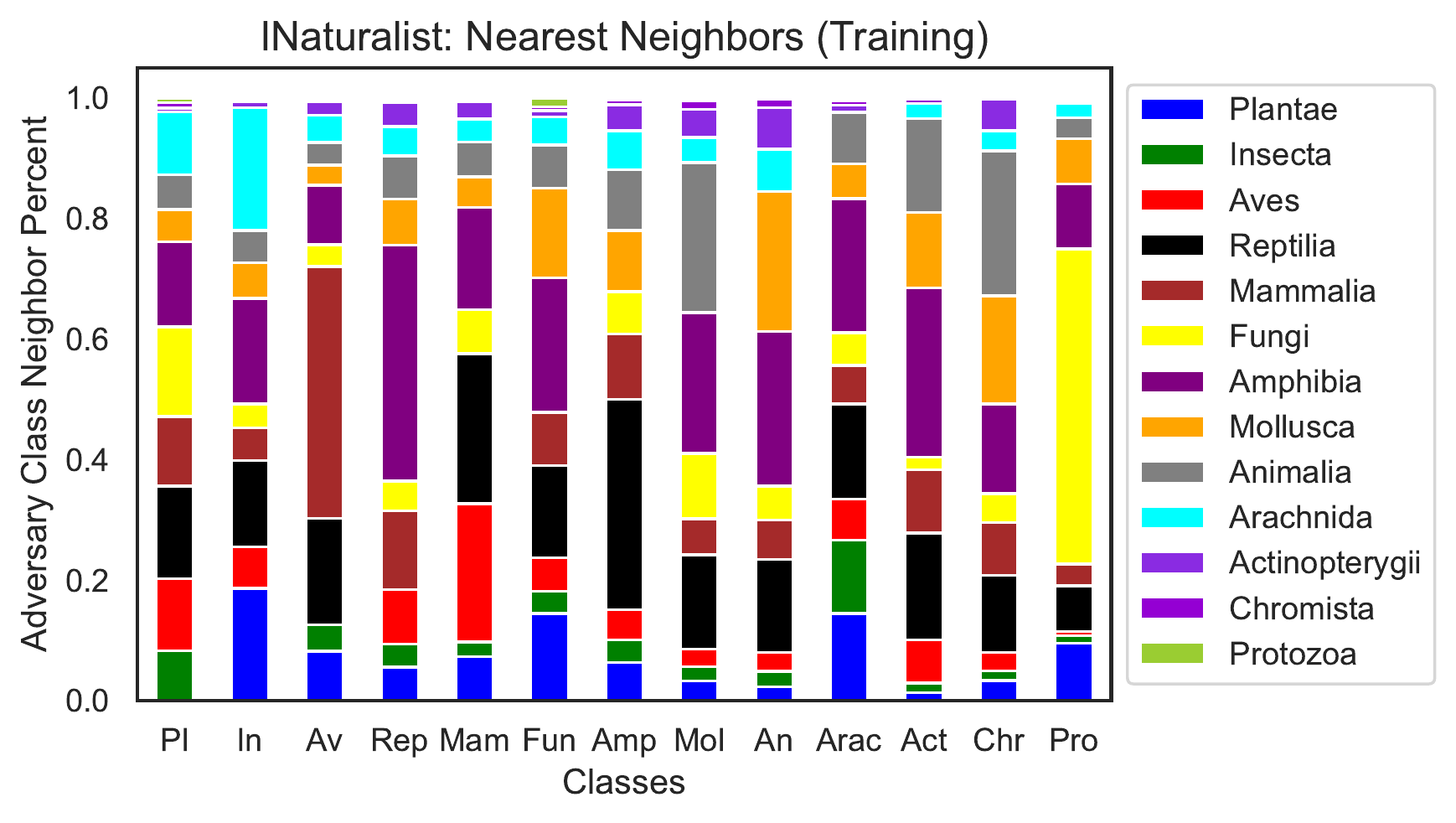}\label{fig:f18}}
   
  \caption{This figure visualizes the relationship between validation error by class and training nearest adversaries by class. This tool can provide a powerful indication of the classes that a model will struggle with during inference; without the costly use of a validation set. Model users and developers can employ the figure on the right as a proxy for the diagram on the left. For example, the training nearest adversary neighbors diagram (b) quickly shows, and the validation set diagram (a) confirms, that the model has the most difficulty (FPs) with: Fungi for the Protozoa class, Mammals for the Aves (birds) class, Amphibia (fish) for the Reptile class, and Reptiles for the Amphibia class.}
  \label{fig_3_nnb}
  \vspace{-0.4cm}
\end{figure*}

\textbf{Use Cases.} For model users, the four categories facilitate the visualization of representative sub-groups within specific classes. When dealing with large datasets, these visualizations can help reduce the need for culling through copious amounts of examples and instead allow model users to focus on a few representative examples: those that are relatively homogeneous in model latent space (safe), those that reside on the decision boundary (border), rare and outlier instances. See Figure~\ref{fig_arch_viz}. 

For imbalanced model developers, the class archetypes can help improve the training process. First, majority class outliers could possibly be mislabeled instances that should be removed. In this case, it may be necessary to examine all of the outlier examples, instead of only the prototype. Second, it can inform potential resampling strategies. For example, safe examples, due to their homogeneity, may be ripe for under-sampling; border and rare instances may be good candidates for over-sampling.

\subsection{Training Set Nearest Neighbors to Gauge Validation Set Error}\label{sec:nnb}

Figure~\ref{fig_3_nnb} visualizes the relationship between validation error and training nearest adversaries by class for a CNN trained with the INaturalist dataset. In the figure, each class is represented with a single bar. The class names are matched with specific colors in the legend. In the figure on the left (a), each color in each bar stands for an adversary class that the model falsely predicts as the reference class. The length of the color bars represent the percentage of total false positives produced by the adversary class. In  Figure~\ref{fig_3_nnb} (a), we show the validation set false positives. 

In contrast, Figure~\ref{fig_3_nnb} (b) shows the percentage of adversary class nearest neighbors for each reference class. By placing these diagrams side by side, we can easily compare how nearest adversaries (on the training set) neatly reproduces the classes that will trigger false positives (in the validation set) and closely matches the percentage of false positives.

This tool can provide a powerful indication of the classes that a model will struggle with during inference when only using training data; without the costly use of a validation set. Model users and developers can employ the figure on the right as a proxy for the diagram on the left. For example, the training nearest adversary neighbors diagram (b) quickly shows, and the validation set diagram (a) confirms, that the model has the most difficulty (FPs) with: Fungi for the Protozoa class, Mammals for the Aves (birds) class, Amphibia (fish) for the reptile class, and reptiles for the Amphibia class.

In order to confirm the ability of training set nearest adversaries to predict the classes that a model will produce more false positives during validation, we measure the difference in the nearest adversary and test FP distributions. We use Kullback Liebler Divergence (KLD) \cite{kullback1951information} to measure the difference in these distributions for five datasets. We also compare our nearest adversary prediction with two other methods: a random distribution and Fisher's Discriminant Ratio. Table~\ref{tab:KLD} shows that our method (NNB) predicts much better than random (by a factor between 1.8 and 34 times better) and compares favorably with another method that measures class overlap, Fisher's Discriminant Ratio.

\begin{figure*}[t!]
   \vspace{-0.6cm}
  \centering
  \subfloat[CIFAR-10: Top-10 Latent Features]{\includegraphics[width=0.49\textwidth]{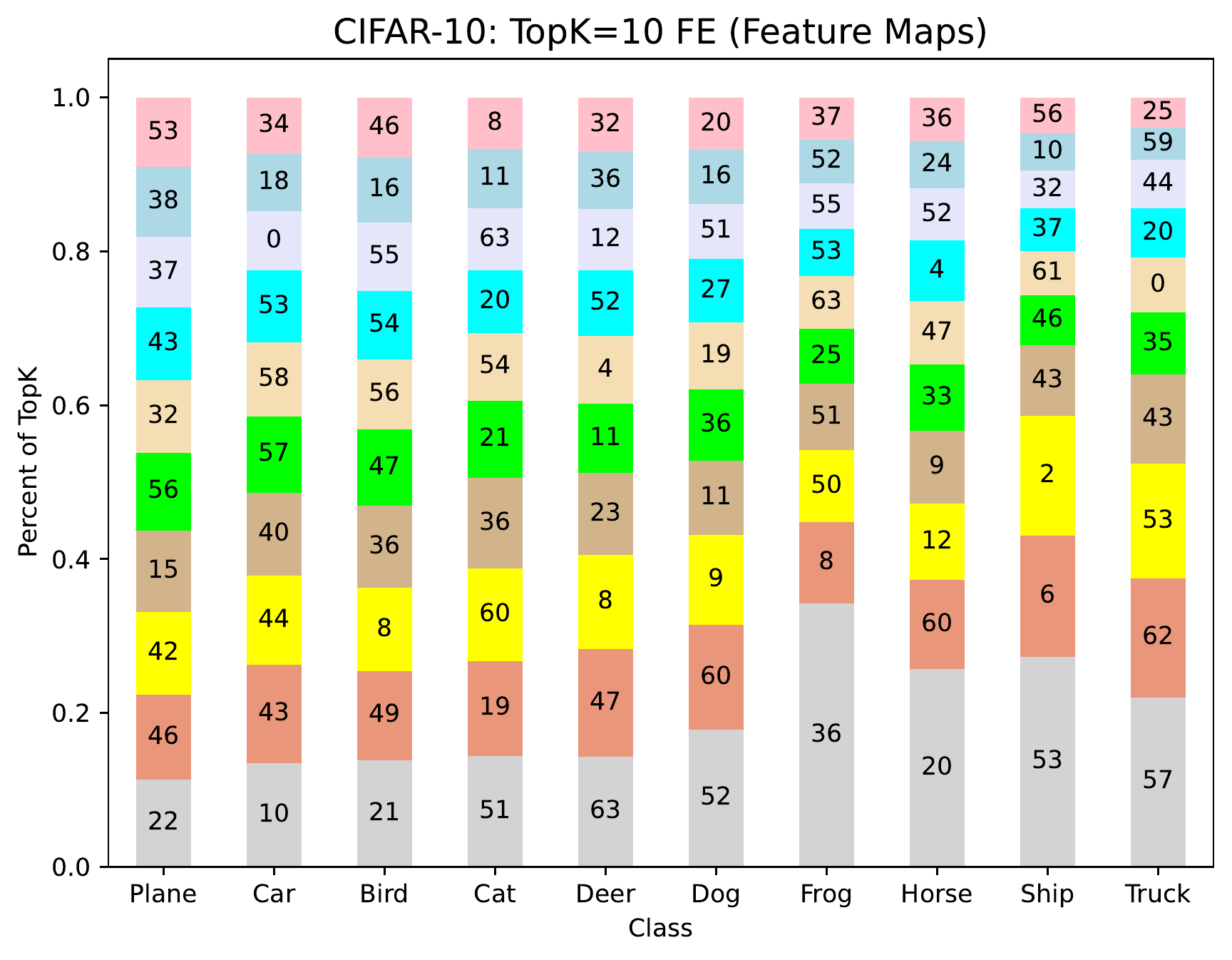}}
  \hfill
  \subfloat[CIFAR-10 (LDAM): Top-10 Latent Features]{\includegraphics[width=0.49\textwidth]{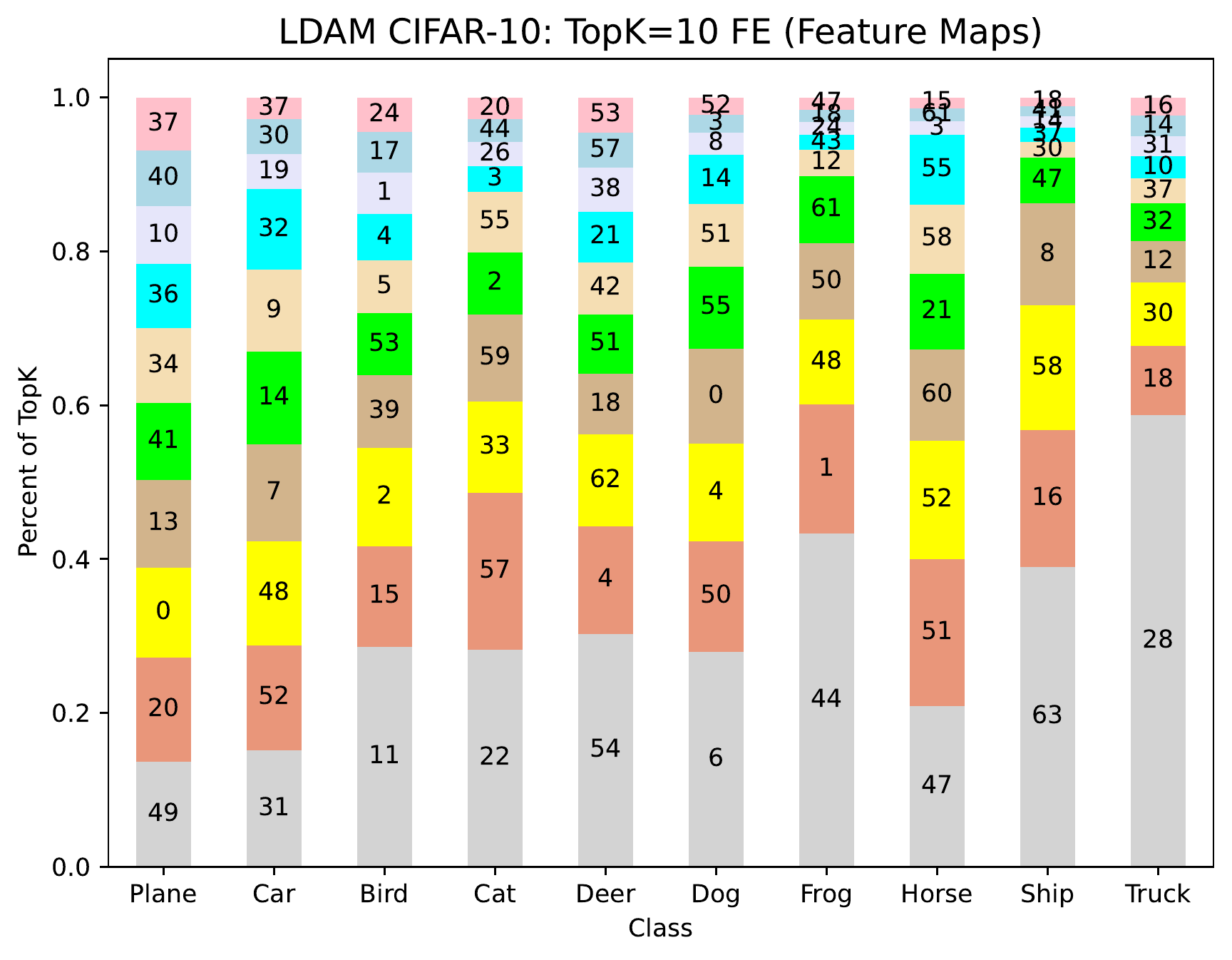}}
   
  \caption{This diagram provides a clear indication of class overlap at the feature map level. It shows the top-K ($K=10$) latent features (FE) used by the model to predict  CIFAR-10 classes. Each of the 10 segments of each bar is color coded, such that gray is the FE with the largest mean magnitude (on the bottom of the bar) and pink is the smallest (top of the bar). In this case, there are a total of 64 feature maps, which correspond to the FE index numbers listed in the bar charts. Each number in the bar chart represents a FE or feature map index. For example, as shown above, whether the model is trained with the LDAM cost sensitive algorithm or simple cross-entropy loss, there is overlap in the top-K latent features for cars and trucks.}
  \label{fig_6_fe_idx}
  \vspace{-0.4cm}
\end{figure*}

This simple tool is useful because it is an indicator of latent feature entanglement. If a model produces a large amount of adversary instances that are in close proximity in the training set to the reference class, then the model will likely have difficulty distinguishing the two close neighbors at validation and test time.

\begin{table}[h!] 
\footnotesize
\caption{KLD of Validation Set False Positives}
\label{tab:KLD}
\centering
\begin{tabular}{ p{1.4cm}p{1.0cm}
p{1.0cm}p{1.0cm}p{1.0cm}}
\toprule

\textbf{Dataset} & \textbf{NNB KLD} &
\textbf{FDR KLD} &
\textbf{Random KLD} & 
\textbf{NNB: \mbox{Random} Factor} \\
\midrule

CIFAR-10 & .5561 & \textbf{.5157} & 4.535 & 8.155 \\
CIFAR-100 & 4.039 & \textbf{1.577} & 7.437 & 1.841\\
Places-10 & .4340 & \textbf{.2942} & 2.821 & 6.500 \\
Places-100 & \textbf{.5919} & 1.327 & 4.191 & 7.081 \\
INaturalist & \textbf{.0577} & .3537 & 1.988 & 34.47 \\
\bottomrule
\end{tabular}
\vspace{-.3cm}
\end{table}

\textbf{Use Cases.} This technique can be a powerful tool for imbalanced data, especially in fields such as chemistry and anomaly detection, where the number of available training examples are precious. In these cases, our method allows model users and imbalanced algorithm developers to gauge the classes that the model will have difficulty with in the validation set, without having to subtract examples to use in the validation process. Therefore, our visualization allows users to reasonably predict the distribution of validation error based solely on the training set. This frees up valuable data that can be dedicated to training and testing (versus validation).

\subsection{Feature Map Overlap}\label{sec:olap}

In the previous section, we visualized class overlap at the instance level. Here, we examine class overlap at the feature embedding (FE) level. FE are scalar values of the output of a CNN's final convolutional layer, after pooling. Higher valued FE indicate CNN feature maps, in the last convolutional layer, that the model views as more important for object classification purposes.

In Figure~\ref{fig_6_fe_idx}, we visualize the ten most significant FE for each class in CIFAR-10 (i.e., the ones with the largest mean magnitudes for each class). Each bar represents a class, as shown on the x-axis. Each of the 10 segments of each bar is color coded, such that gray is the FE with the largest mean magnitude (on the bottom of the bar) and pink is the smallest (top of the bar). Each color coded segment of a bar contains a number, which is the index of a FE / FM. For this model, there are 64 FE / FM (0 to 63). The relative size of each segment (y-axis) shows the percentage that each FE magnitude makes up of the top-10 FE magnitudes (i.e., they all sum to 1).

Therefore, the chart shows the most important latent features (feature maps) that a CNN uses for each class to make its class decision. Because the FE indices are shown for each class, they can be compared between classes to identify latent feature overlap. In other words, it shows latent feature maps that the model uses to distinguish multiple classes.

\begin{figure}[h!]
   \vspace{-0.2cm}
  \centering
  \subfloat[CE: Trucks FPs]{\includegraphics[width=0.24\textwidth]{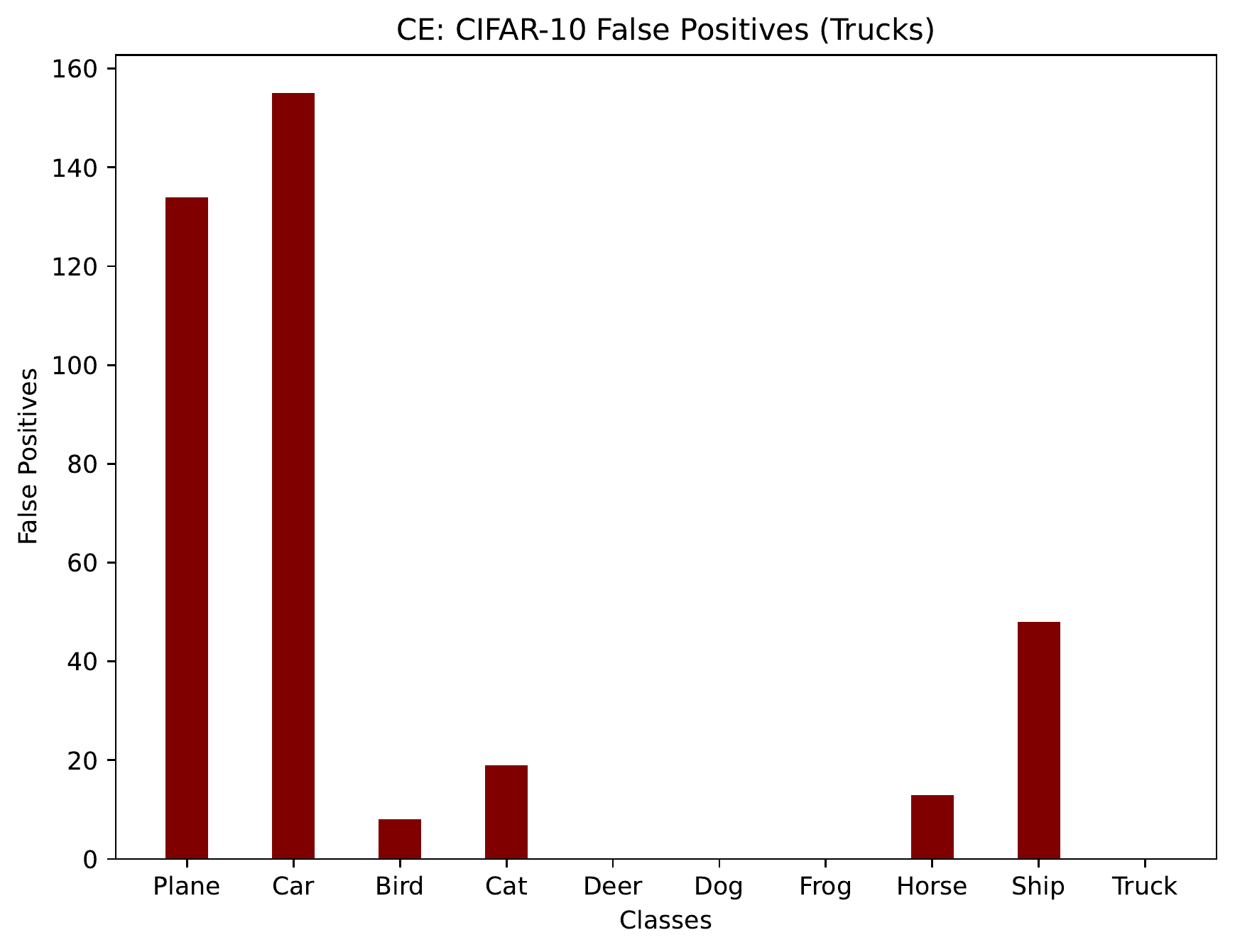}}
  \hfill
  \subfloat[LDAM: Trucks FPs]{\includegraphics[width=0.24\textwidth]{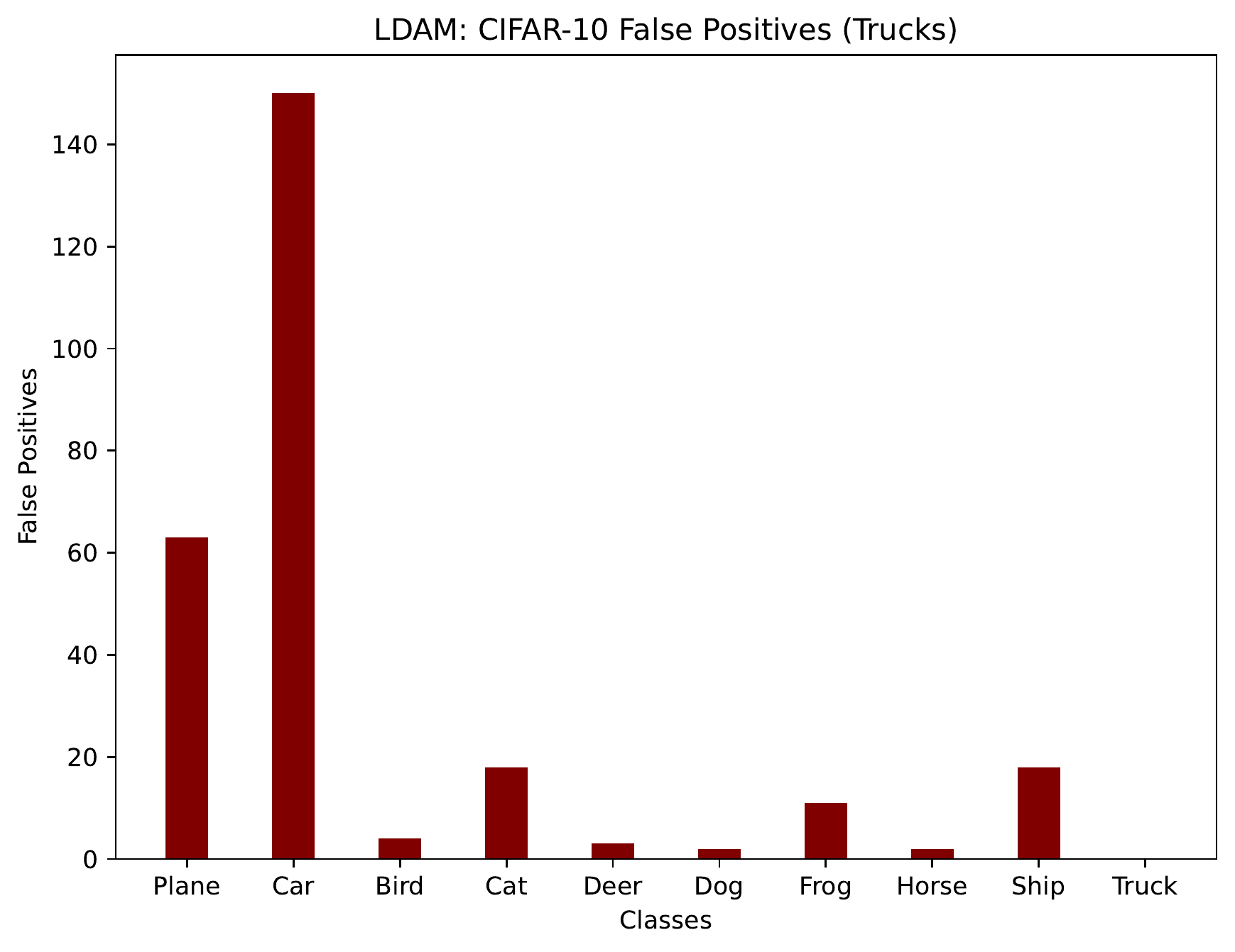}}
   
  \caption{This diagram shows the false positives for trucks for CNNs trained with cross-entropy loss (CE) and LDAM.}
  \label{fig_fe_fp}
  \vspace{-0.2cm}
\end{figure}

For example, in Figure~\ref{fig_6_fe_idx} (a), we can see that trucks and cars contain five common FE in their top-10 most important FE (i.e., FE indices 57, 53, 43, 0 and 44). In contrast, trucks and planes share only 2 top-10 FE (FE indices 43 and 53). Trucks are the class with the fewest number of training examples, with planes the most, and cars the next largest. For trucks, the two classes that produce the most false positives at validation time are cars and planes, respectively (see Figure~\ref{fig_fe_fp} (a)). This chart implies that the large number of FPs produced for planes and cars may be due to two different factors. In the case of planes, it seems to be due to numerical differences in the number of training examples because of the low FE overlap; whereas, in the case of cars, it appears to be due to FE overlap.

\begin{figure*}[b!]
   \vspace{-0.2cm}
  \centering
  \subfloat[Truck Textures (Minority Class)]{\includegraphics[width=0.33\textwidth]{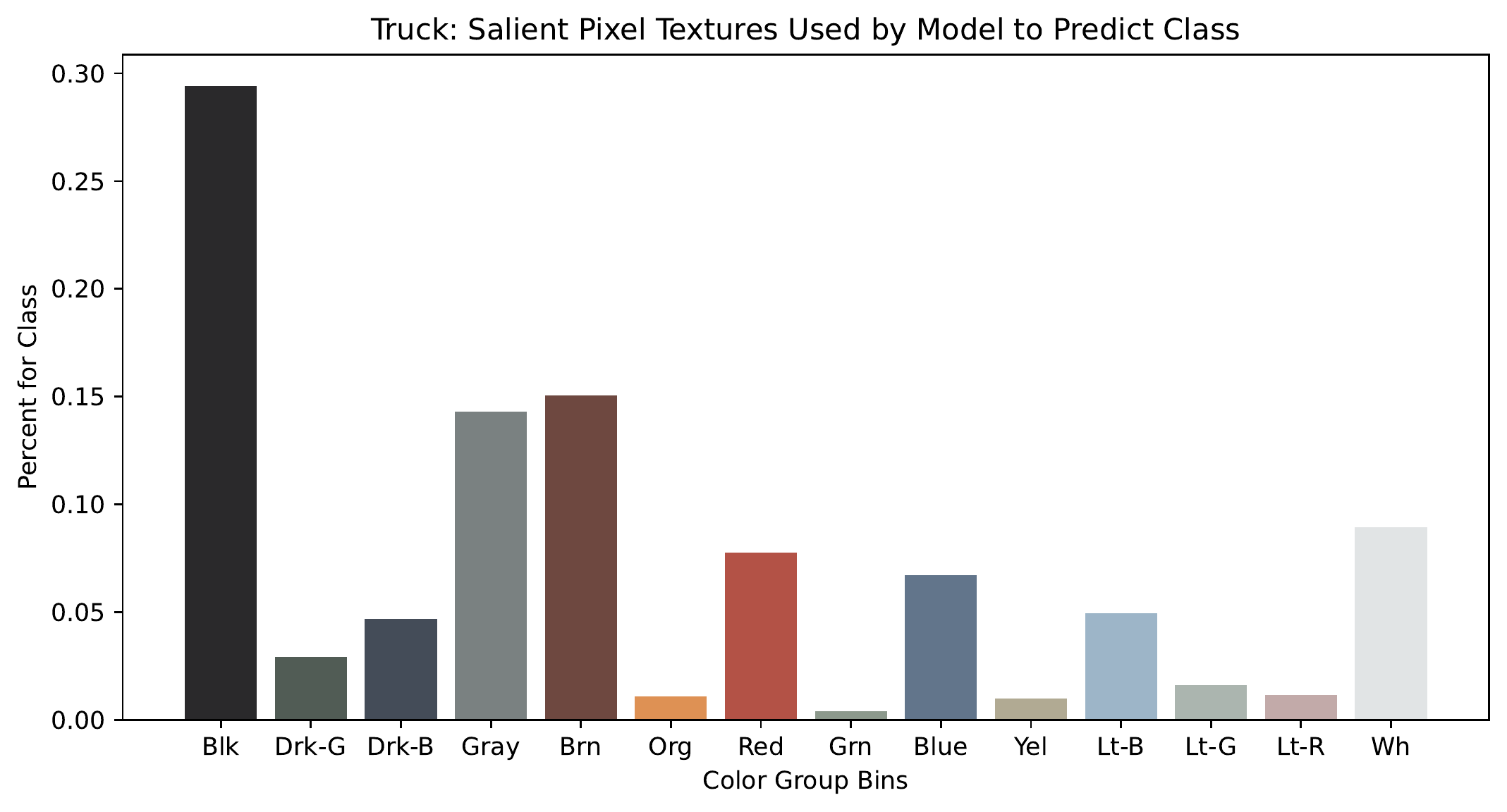}}
  \hfill
  \subfloat[Auto Textures (Overlap)]{\includegraphics[width=0.33\textwidth]{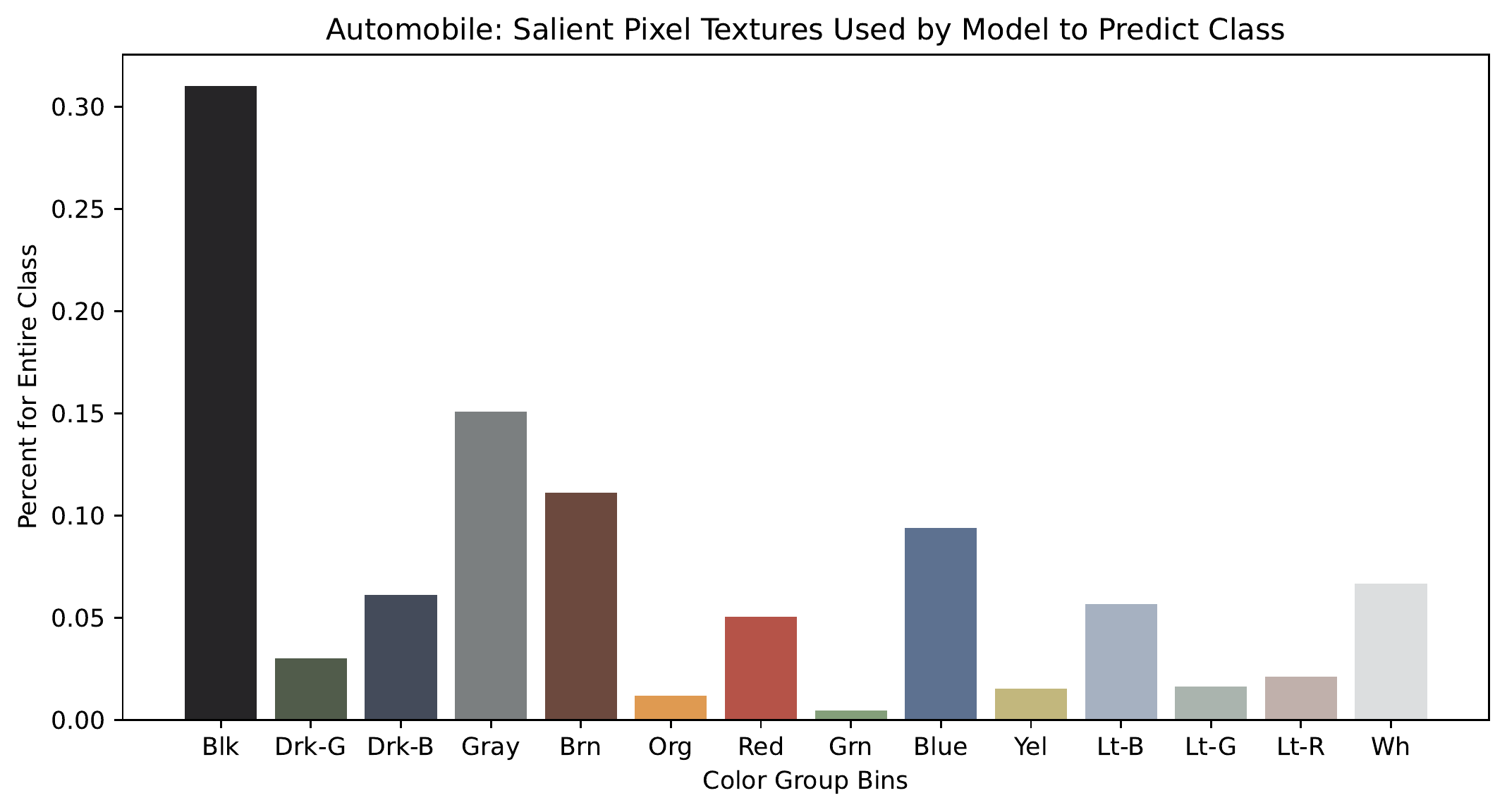}}
  \hfill
  \subfloat[Plane Textures (Majority Class)]{\includegraphics[width=0.33\textwidth]{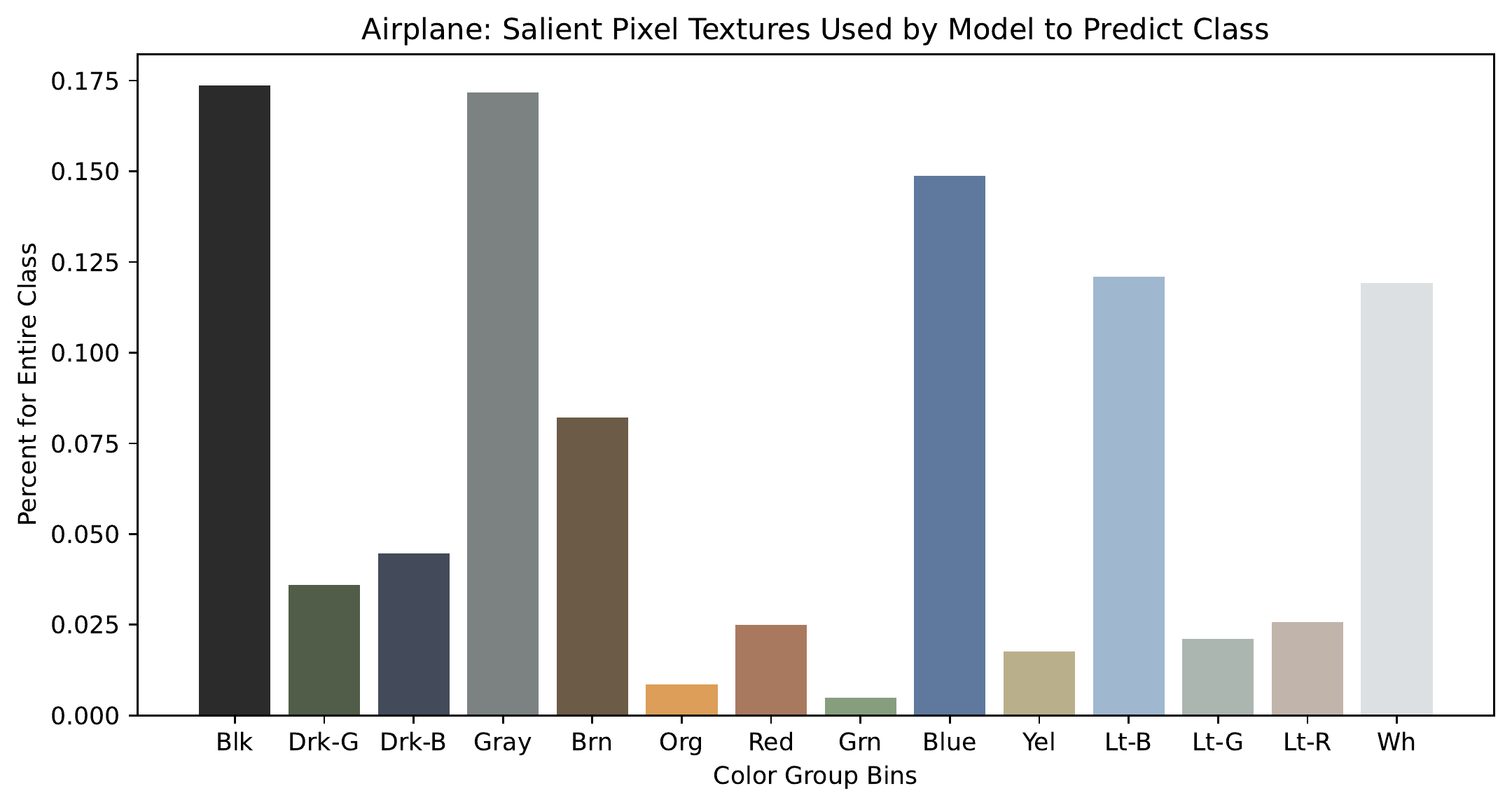}}
 
  \caption{This diagram shows the top 10\% of color groups for specific classes based on gradient saliency tracing. The classes are drawn from CIFAR-10. In the case of autos and trucks, black ($\approx30\%$) and gray ($\approx15\%$) are the 2 most common colors. Since all cars and trucks have (black) tires, the presence of this color is not surprising. Even though the number of samples is vastly different between cars and trucks (60:1 imbalance), the overall proportion of colors is very similar, which tracks the feature embedding space overlap that we previously observed for these 2 classes (model feature entanglement). In the case of planes, black and gray are still important ($\approx17\%$ each); however, there is a much larger percentage of blue, light blue, and white ($\approx12.5\%$ each), due to the greater presence of blue sky and white clouds (background). In contrast, white is salient only $\approx5\%$ of the time for cars and trucks.}
  \label{fig_6_fe_texture}
  \vspace{-0.4cm}
\end{figure*}

We can further explore this hypothesis by examining how a cost-sensitive algorithm, LDAM, which focuses on the numerical difference of training instances (and not features) behaves in the face of class overlap.

In Figure~\ref{fig_6_fe_idx}, the figure on the left (a) is trained with cross-entropy loss and the figure on the right (b) is trained with a popular cost-sensitive method used in imbalanced learning, LDAM. 
Interestingly, in the figure on the right (b), where a CNN is trained with a cost-sensitive method, there is still five FE that are shared in common between the truck and car classes. In fact, if we view figures (a) and (b) of Figure~\ref{fig_fe_fp}, we can see that LDAM reduces false positives for the plane class but does not have a large impact on the automobile class, which is likely because it is geared toward addressing instance numerical differences and not latent feature overlap.  Thus, although the cost-sensitive method may have addressed the class imbalance, in part, it does not appear to have completely addressed feature overlap.

\textbf{Use Cases.} This visualization can provide vital clues about where a CNN classifier may break-down. The cause of FPs may not always be solely due to class imbalance. Other factors, such as a model's entanglement of latent features, may be at stake. In these situations, imbalanced learning algorithm developers may want to consider techniques that address feature entanglement, instead of solely class numerical imbalance. For example, it may be possible to design cost-sensitive loss functions that assign a greater cost to FE overlap based on FE index commonality between classes. 

This visualization may also serve as a helpful tool when used for comparative purposes. It can be used as a comparison between baseline loss functions and cost-sensitive algorithms. The visualization can help imbalanced learning algorithm developers decide if, for example, cost-sensitive techniques are addressing only class imbalance or, additionally, if their methods improve feature entanglement in latent space.

\subsection{Feature Density}\label{sec:den}

This visualization combines both FE overlap with density. It zooms in on a particular class and compares the density of its top-K FE indices with the same top-K FE indices in adversary classes. In other words, given a set of top-K FE indices for a reference class, it counts the number of instances in an adversary class where the reference class top-K FE indices are also in the top-K FE indices of an adversary class. This provides an indication of the density of reference class top FE indices compared to adversary class top FE indices.

\begin{figure}[h!]
   \vspace{-0.3cm}
  \centering
  \subfloat[CE: Trucks Latent Feature Density]{\includegraphics[width=0.47\textwidth]{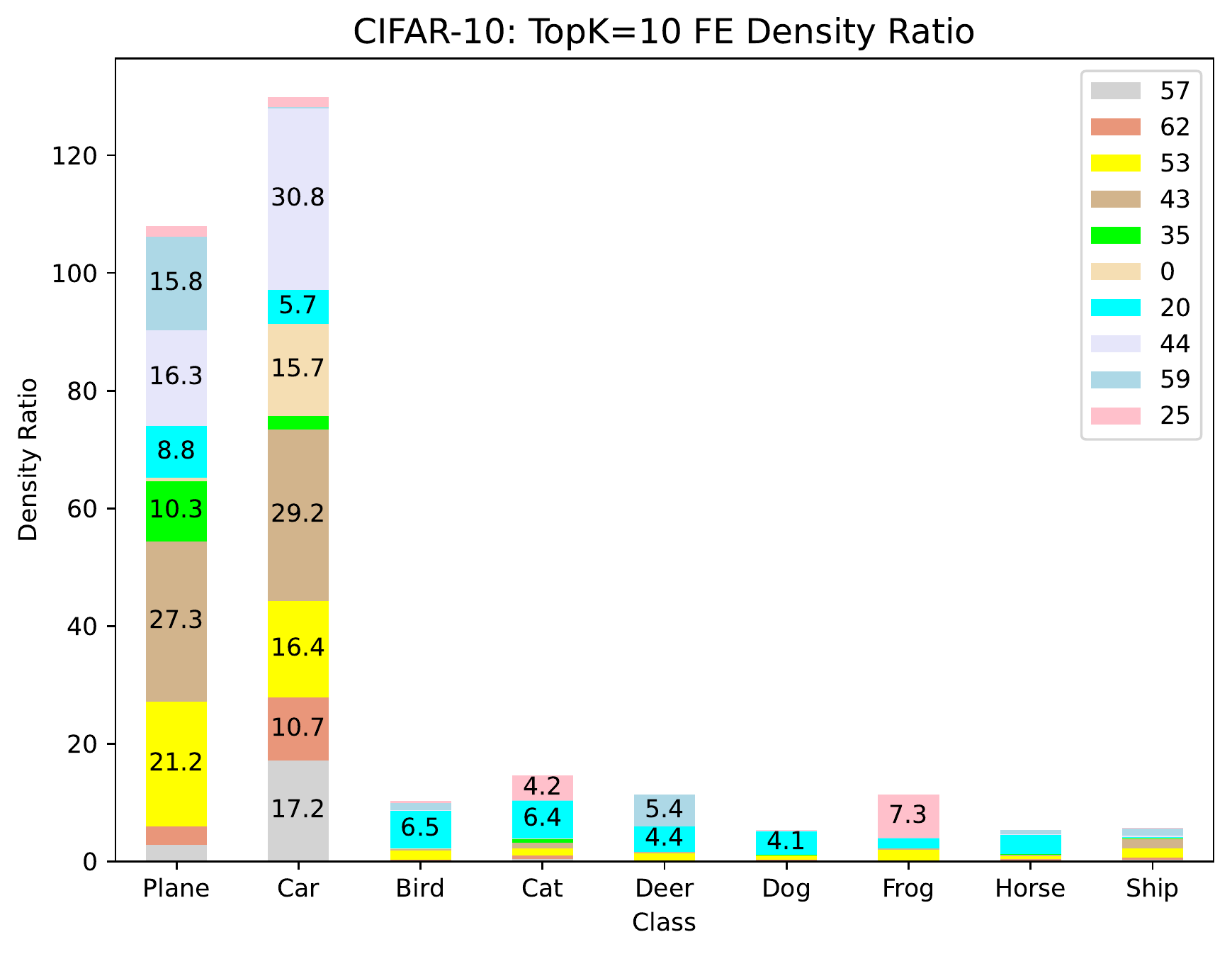}}
   
  \caption{This diagram shows the importance of latent feature density. The ratio of the top-10 FE for the minority class (trucks) relative to the same FE in each adversary class is illustrated. The legend shows the feature map index number for the reference class (trucks in CIFAR-10) and the bars show the density level for the respective FE.   The density of minority class top-10 features is far outweighed by the same majority class FE.}
  \label{fig_6_fe_density}
  \vspace{-0.5cm}
\end{figure}

Figure~\ref{fig_6_fe_density} focuses in on a single class: the extreme minority class in CIFAR-10 (trucks). It compares the density ratio of each of the top-10 FE in trucks (a reference class or class of interest) to each of the other classes in the dataset (each an adversary class). Each adversary class is listed on the x-axis, along with a color coded bar. The colors in each bar correspond to the top-10 most important FE indices in the reference class (trucks), which are shown in the legend. The numbers in the bar segments are the density ratio of a particular FE. For example, the number 30.8 in the cars bar means that FE index 44 appears 30.8 times more frequently in latent space as a top-10 FE magnitude in the cars versus trucks class. The ratio is arrived at by dividing the number of instances that an individual FE is a top-10 feature for cars versus trucks. 

In Figure~\ref{fig_6_fe_density}, we can see that the two classes that have the highest density of similar features to trucks are planes and cars. In Figure~\ref{fig_fe_fp} (a), we can similarly see that they are the same two classes that produce the most false positives with respect to trucks in the validation set.

\begin{figure*}[!t]
   \vspace{-0.5cm}
  \centering
  \subfloat[Safe]{\includegraphics[width=0.25\textwidth]{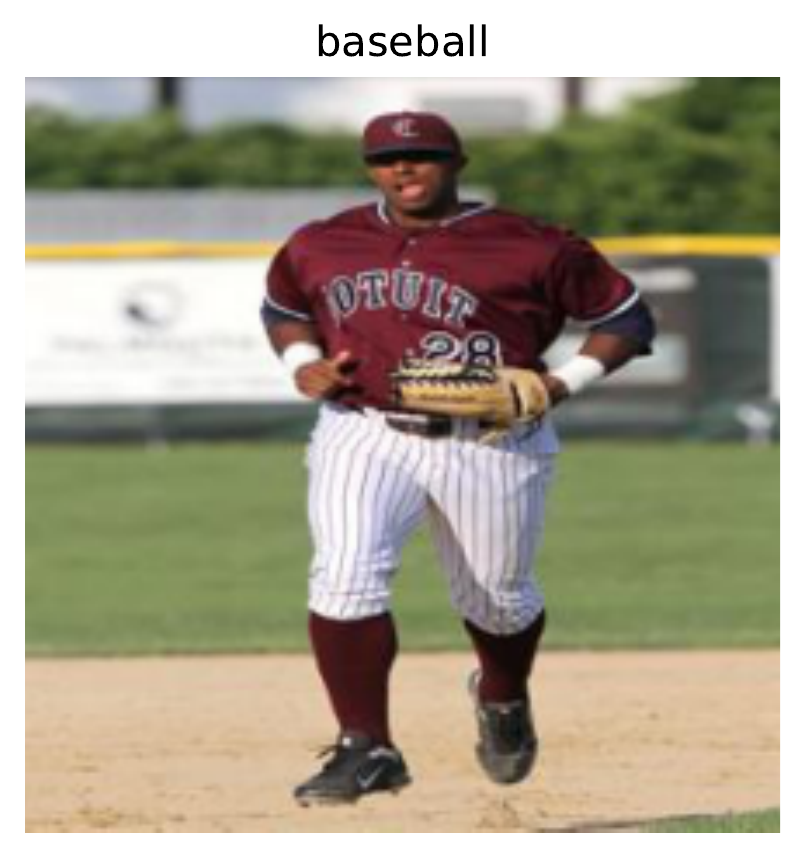}}
  \subfloat[Border]{\includegraphics[width=0.25\textwidth]{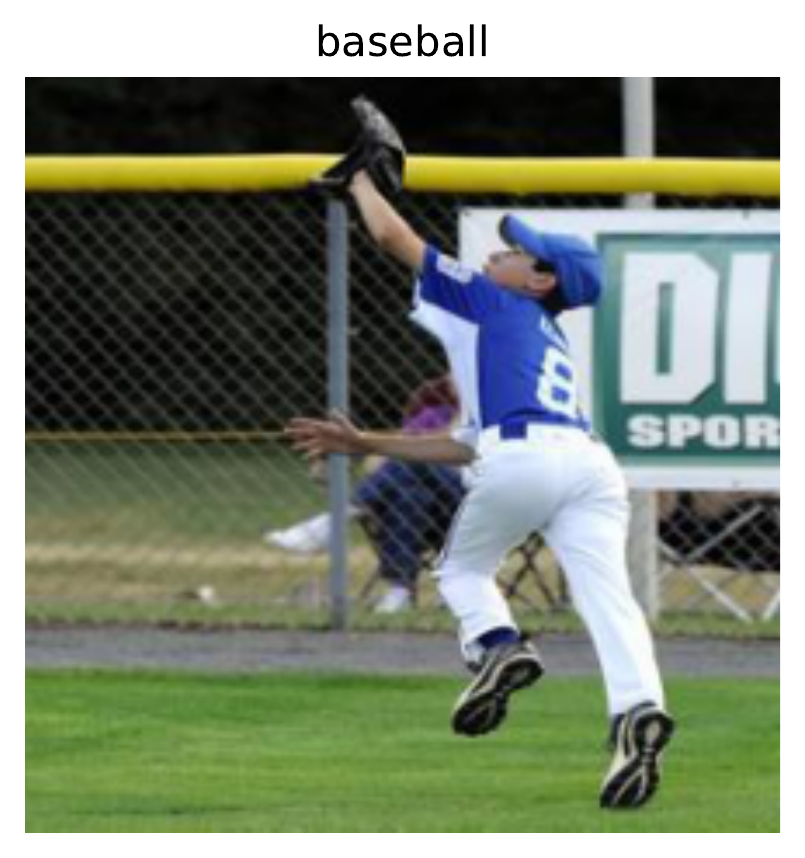}}
  \hfill
  \subfloat[Safe]{\includegraphics[width=0.25\textwidth]{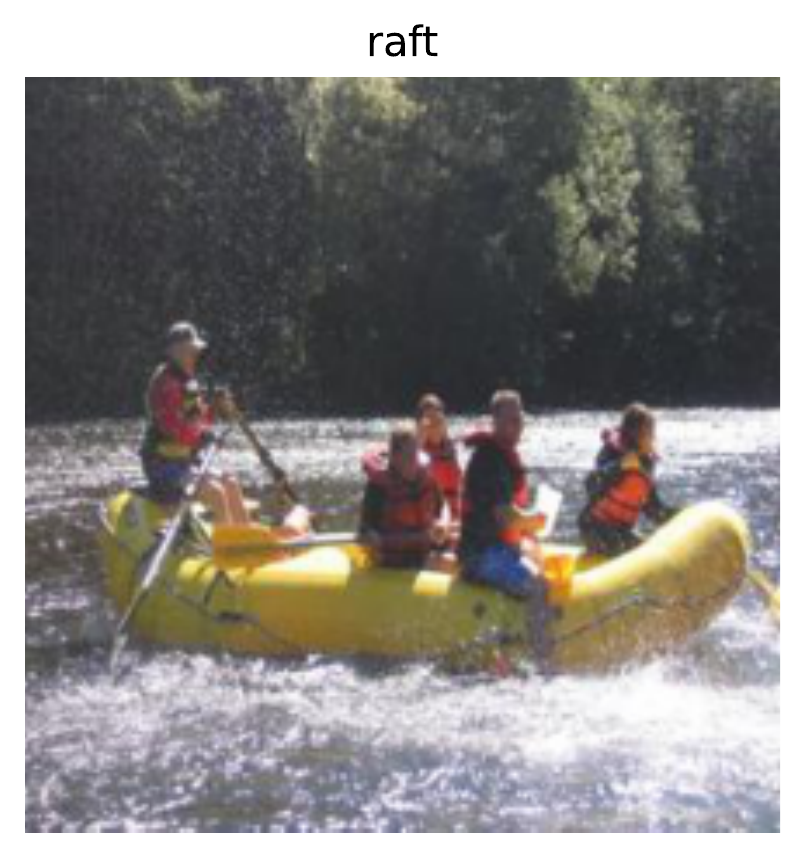}}
 \hfill
  \subfloat[Border]{\includegraphics[width=0.25\textwidth]{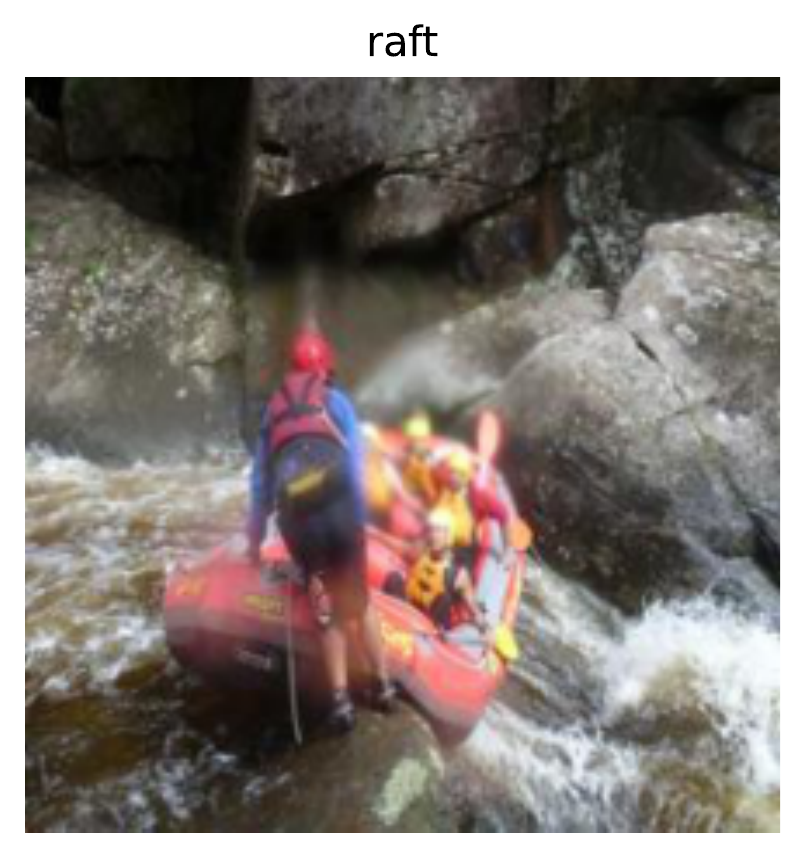}}
  \hfill
  \subfloat[Baseball Field Textures]{\includegraphics[width=0.49\textwidth]{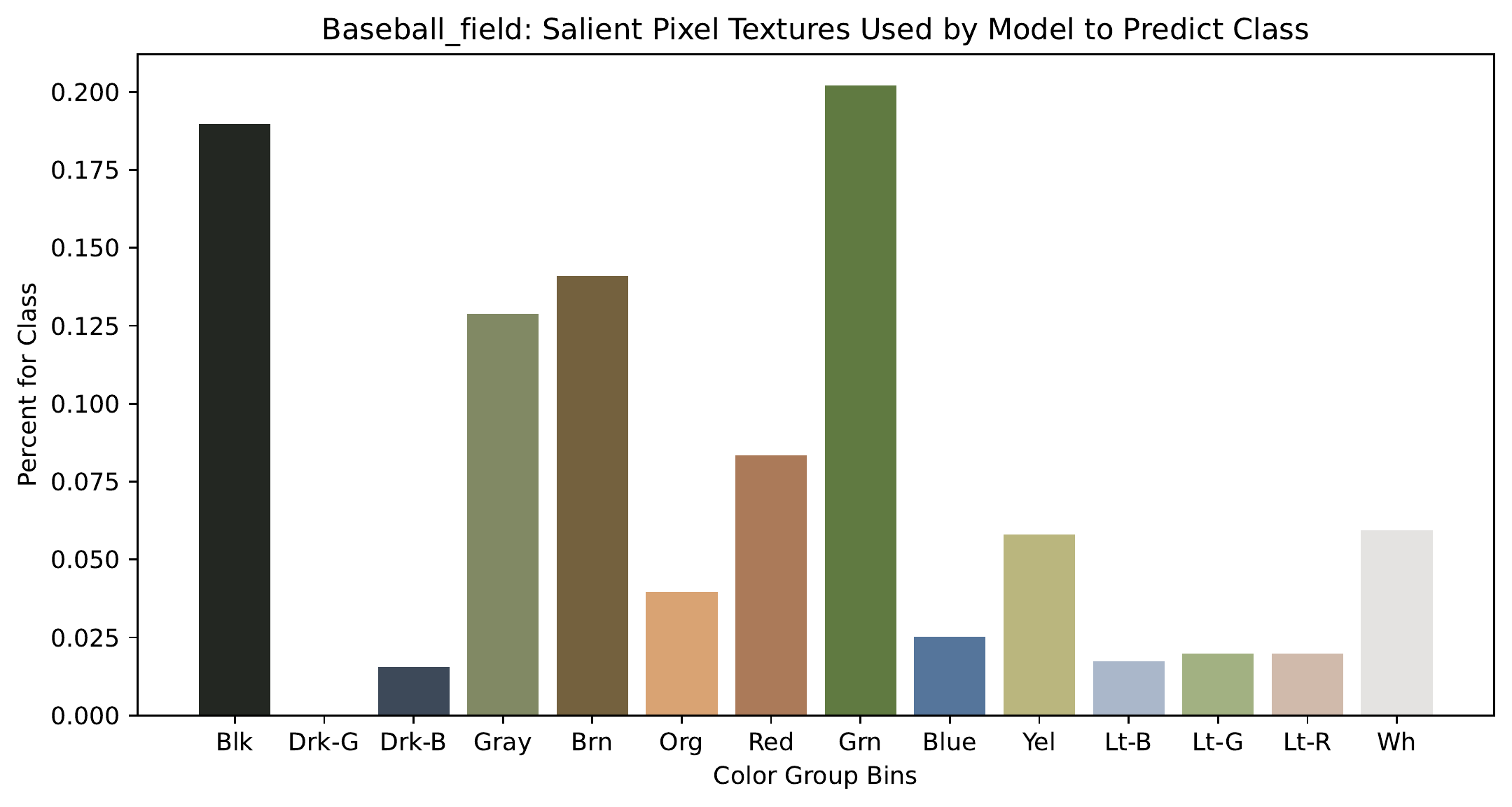}}
  \hfill
  \subfloat[Raft Textures]{\includegraphics[width=0.49\textwidth]{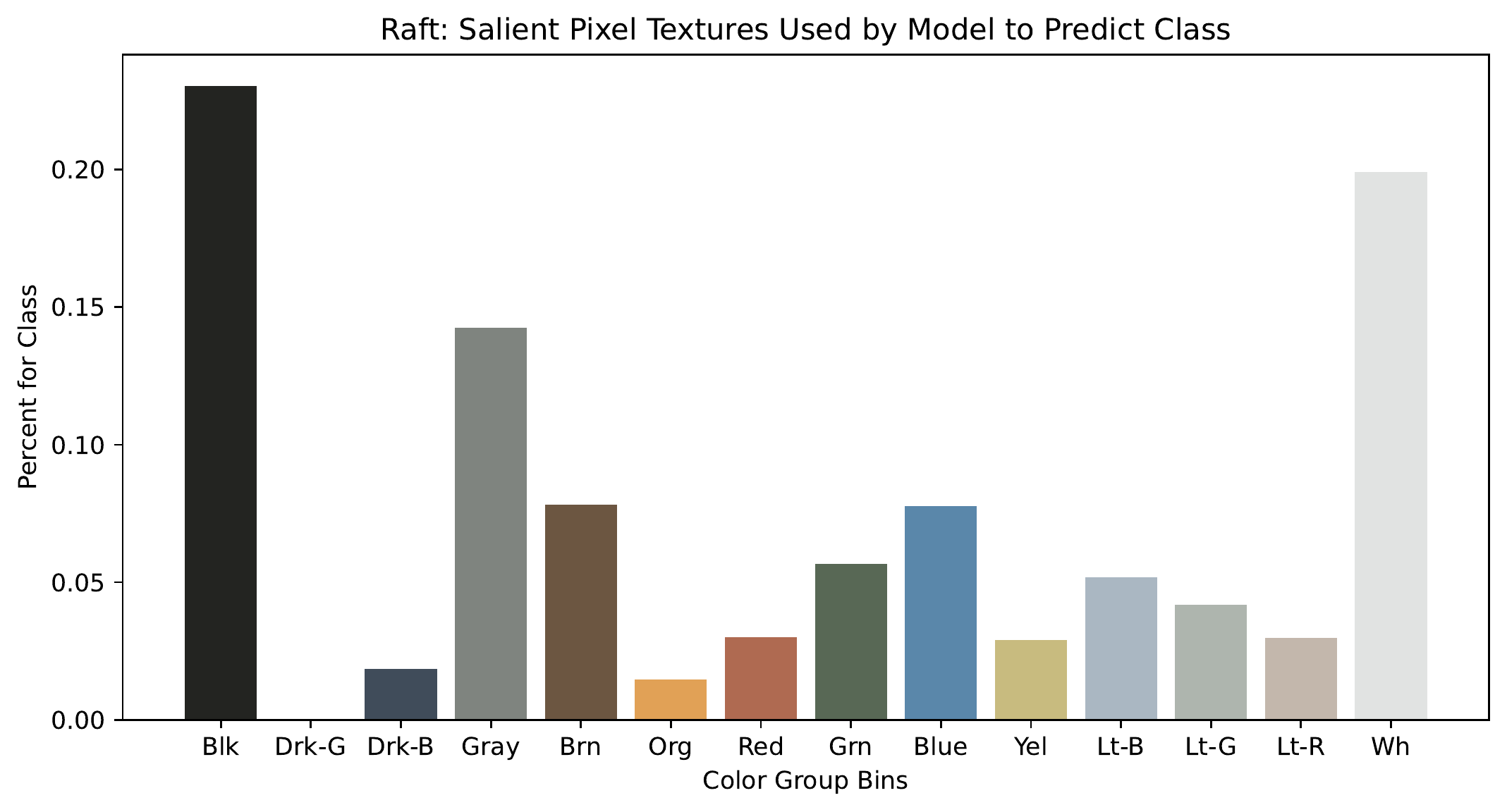}}
  
  \caption{This diagram shows the most salient colors for a majority (baseball field) and minority class (raft), along with archetypical images drawn from the safe and border categories and a CNN trained on the Places-100 dataset. In the case of a baseball field, the model preferences green, brown and gray; whereas for rafts, white is more prevalent (likely due to white rapids) and brown and green are less emphasized. This type of information may be relevant for purposes of oversampling techniques in pixel space. By determining the colors that the model preferences, it may be possible to modify the colors via augmentation to train the model to preference other colors.}
  \label{fig_texture_places}
  \vspace{-0.4cm}
\end{figure*}

\textbf{Use Cases.} This visualization may prove useful to designers of resampling techniques for imbalanced data. For example, it provides clues as to both the specific latent features and the relative densities of those features. Oversampling methods such as EOS \cite{dablain2022efficient}, which resample in feature embedding space, may serve as a good base algorithm that can be specifically tailored to perform oversampling on specific FE. It may also inform the amount of samples that should be created. Simply equalizing the number of class instances, without evaluating the impact of specific features on class imbalance, may not provide optimal accuracy improvements.

\subsection{Colors that Define Classes}\label{sec:text_exp}

This visualization can be used to identify the color bands that are most prevalent in a data class. As an illustration, Figure~\ref{fig_6_fe_texture} shows the top 10\% of color group textures for the truck, auto and plane classes in CIFAR-10. In the case of autos and trucks, black (30\%) and gray (15\%) are the 2 most common colors. Since all cars and trucks have (black) tires, the presence of this color is not surprising. Even though the number of samples is vastly different between cars and trucks (60:1 imbalance), the overall proportion of color bands is very similar, which tracks the FE space overlap that we previously observed for these 2 classes (model feature entanglement). In the case of planes, black and gray are still important (17\% each); however, there is a much larger percentage of blue, light blue, and white (12.5\% each), due to the greater presence of blue sky and white clouds (background). In contrast, white is salient only 5\% of the time for cars and trucks.

Additionally, Figure~\ref{fig_texture_places} shows safe and border prototypes for a baseball field (majority class) and rafts (minority class) from a CNN trained on the Places-100 dataset with cross-entropy loss. For the baseball field, the most salient colors used by the model to detect class instances are black, green and brown. In the safe prototype, we can see black leggings on the player's uniform, green grass and a brown infield. In the border prototype, we can see a black background (over the fence),  player black shoes, and green grass. In the case of rafts, green and brown are not as prevalent in the model's top 10\% most salient pixels. Instead, white (white water rapids) and the black background are more important.

\textbf{Use Cases.} Users of CNNs trained on imbalanced data may use this visualization to better understand the major color bands that are prevalent across a class. When combined with class prototype visualization, it can also provide intuition into whether a classifier is using background colors (e.g., blue sky or clouds) to discern a class. For imbalanced learning algorithm developers, it can suggest specific pixel color groups that may be over- or under-sampled at the front-end of image processing to improve classifier accuracy.

\section{Conclusion}
We present a framework that can be used by both model users and algorithm developers to better understand and improve CNNs that are trained with imbalanced data. Because modern neural networks depend on large quantities of \textit{data} to achieve high accuracy, understanding how these models use complex data is critical. Our framework enables model users and developers to visualize and better understand class overlap at the instance and latent feature levels, the density of latent features, and the role that color plays in CNN class decisions.

\bibliographystyle{IEEEtran}
\bibliography{references}

\end{document}